\documentclass[runningheads]{llncs}
\usepackage{graphicx}
\usepackage{amsmath,amssymb} 
\usepackage{color}

\usepackage{algorithm}      
\usepackage[noend]{algpseudocode} 
\usepackage{caption}
\usepackage{subfig}
\usepackage{esvect}         
\usepackage[toc]{appendix}
\usepackage{wrapfig}        
\usepackage{soul}           
\usepackage{afterpage}      
\usepackage{tabularx}

\usepackage[utf8]{inputenc}
\usepackage{bm}
\usepackage[section]{placeins}

\usepackage[width=122mm,left=12mm,paperwidth=146mm,height=193mm,top=12mm,paperheight=217mm]{geometry}

\newcommand\blfootnote[1]{%
  \begingroup
  \renewcommand\thefootnote{}\footnote{#1}%
  \addtocounter{footnote}{-1}%
  \endgroup
}

\usepackage{array}
\newcolumntype{L}[1]{>{\raggedright\let\newline\\\arraybackslash}m{#1}}
\newcolumntype{C}[1]{>{\centering\let\newline\\\arraybackslash}m{#1}}
\newcolumntype{R}[1]{>{\raggedleft\let\newline\\\arraybackslash}m{#1}} 

\usepackage{xspace}
\makeatletter
\DeclareRobustCommand\onedot{\futurelet\@let@token\@onedot}
\def\@onedot{\ifx\@let@token.\else.\null\fi\xspace}

\def\ie{\emph{i.e}\onedot}

\def\etal{\emph{et al}\onedot}
\makeatother

\begin{document}

\pagestyle{headings}
\mainmatter

\title{Escaping from Collapsing Modes in a Constrained Space} 

\titlerunning{Escaping from Collapsing Modes in a Constrained Space}

\authorrunning{C. Chang, C. Lin, C. Lee, D. Juan, W. Wei and H. Chen}

\author{
    Chia-Che Chang*$^1$ \and Chieh Hubert Lin*$^1$ \and Che-Rung Lee$^1$ \\
    Da-Cheng Juan$^2$ \and Wei Wei$^3$ \and Hwann-Tzong Chen$^1$
}

\institute{$^1$Department of Computer Science,\\
    National Tsing Hua University\\
    \email{\{chang810249,hubert052702,cherung\}@gmail.com},\\
    \email{htchen@cs.nthu.edu.tw},\\
    $^2$Google AI, Mountain View, CA, USA \\
    \email{dacheng@google.com} \\
    $^3$Google Cloud AI, Mountain View, CA, USA \\
    \email{wewei@google.com}
}

\maketitle

\blfootnote{* Indicates equal contribution.}

\begin{abstract}

    Generative adversarial networks (GANs) often suffer from unpredictable mode-collapsing during training. We study the issue of mode collapse of Boundary Equilibrium Generative Adversarial Network (BEGAN), which is one of the state-of-the-art generative models. Despite its potential of generating high-quality images, we find that BEGAN tends to collapse at some modes after a period of training. We propose a new model, called \emph{BEGAN with a Constrained Space} (BEGAN-CS), which includes a latent-space constraint in the loss function. We show that BEGAN-CS can significantly improve training stability and suppress mode collapse without either increasing the model complexity or degrading the image quality. Further, we visualize the distribution of latent vectors to elucidate the effect of latent-space constraint. The experimental results show that our method has additional advantages of being able to train on small datasets and to generate images similar to a given real image yet with variations of designated attributes on-the-fly.
    
\end{abstract}

\section{Introduction}
    
    The main goal of this paper is to provide new insights into the problem of mode collapse in training Generative Adversarial Networks (GANs)~\cite{GAN}. GANs have shown great potential in generating new data based on real samples and have been applied to various vision tasks~\cite{BousmalisSDEK17,DaiFUL17,GwakCGCS17,LedigTHCCAATTWS17,LiLY017,ShrivastavaPTSW17,SoulySS17,TzengHSD17}. Our study points out a simple but effective approach that can be used to improve the stability of training GANs for generating high-quality images with respect to disentangled representations.  
    
    GANs comprise two core components: generator $G$ and discriminator $D$. The two components are optimized with respect to two spaces. One is the latent space $Z$ for the generator, and the other is the data space $X$ associated with a real data distribution $p_{\mathrm{real}}(x)$ for training data $x \in X$. The objective of the generator is to find a mapping $G: Z \rightarrow X$ that maximizes the probability of the discriminator mistakenly accepting a generated image $G(z), z \in Z$ as from $p_{\mathrm{real}}(x)$. On the contrary, the discriminator's objective is to distinguish whether any given $x \in X$ belongs to $p_{\mathrm{real}}(x)$. During training, the generator only learns from the information provided by the discriminator, and aims to estimate a good mapping such that $p_{\mathrm{model}}(G(z))$ is similar to $p_{\mathrm{real}}(x)$.
    
    Compared with auto-encoders~\cite{AutoEncoder}, GANs can generate sharper images owing to the adversarial loss. However, a downside of adopting the adversarial loss is that it makes the training of GANs unstable. The performance is strongly dependent on hyper-parameters selection, and the generated images tend to have weaker structural coherence.
   
    Boundary Equilibrium Generative Adversarial Network (BEGAN)~\cite{BEGAN} introduced by Berthelot \etal suggests several modifications on the architecture and loss designs, which significantly improve the quality of generated images and the training stability. Another contribution of BEGAN is providing an approximation of convergence for the class of energy-based GANs.
    
    Despite the promising improvements of BEGAN, we empirically observe that BEGAN still unavoidably runs into mode collapses after certain epochs of training. In the meanwhile, neither the approximation of convergence nor the loss functions of BEGAN is able to detect the sudden mode collapses. In our experiments, the exact time when mode collapsing happens is highly related to target image resolution and dataset size. In addition to the typical drawbacks of mode collapsing, this unpredictable behavior also makes BEGAN's intended contribution to providing ``global measure of convergence'' incomplete.

    \subsection{Contributions}
    
    We propose a new constraint loss toward addressing the mode collapsing problem. We find that the mode-collapsing problem is suppressed after adding the constraint loss. This new loss term does not increase model complexity and is computationally low-cost. Furthermore, it does not introduce any trade-off regarding image quality and diversity. 
    The proposed model is called \emph{BEGAN with a Constrained Space} (BEGAN-CS).

    We visualize the latent vectors produced in training phase using Principal Component Analysis (PCA)~\cite{PCA}. In section \ref{subsection:PCA-analysis}, we analyze the effect of the constraint loss and explain why this loss term makes training process stable.
    
    Since BEGAN-CS is more stable during training, it performs consistently well even when the size of training dataset is ten-times smaller than the normal setting, in which BEGAN fails to obtain acceptable results. In section \ref{subsection:better-convergence-on-small-dataset}, our experiment shows that the proposed BEGAN-CS can eventually converge to a better state, while BEGAN ends up at mode collapsing in an early stage.
    
    We further discover that BEGAN is able to learn strong and high-quality disentangled representations in an unsupervised setting. The learned disentangled representations could be used to modify the underlying attributes of generated images. In the meanwhile, owing to the constraint loss, BEGAN-CS can accomplish approximation $Enc(x^*) \simeq z^*$ on-the-fly for any given real image $x^*$, where $G(z^*)$ is an approximate image to $x^*$ under the fixed generator weights. 
    By leveraging the $z^*$ approximation and the disentangled representations, BEGAN-CS can generate on the fly a set of images conditioning on a real image $x^*$. The generated images are visually similar to the given real image and are able to exhibit the adjustable disentangled attributes.

\section{Related Work} 
    
    Deep Convolutional Generative Adversarial Network (DCGAN)~\cite{DCGAN} improves the original GAN~\cite{GAN} by employing a convolutional architecture to achieve better stability of training and enhanced quality of generated images. Salimans~\etal further present several practical techniques for training GANs~\cite{SalimansGZCRCC16}. Nevertheless, avoiding mode collapsing while keeping the quality of generated images is still a challenging issue in practice. 
     
    Energy-Based Generative Adversarial Network (EBGAN)~\cite{EBGAN} introduces another perspective for formulating GANs. EBGAN implements the discriminator as an auto-encoder with per-pixel error. Boundary Equilibrium Generative Adversarial Network (BEGAN)~\cite{BEGAN} shares the same discriminator setting as EBGAN and makes several improvements on the designs of architecture and loss function. One of BEGAN's core contributions is introducing the equilibrium concept, which balances the power between the generator and the discriminator. With these improvements, BEGAN provides fast and stable training convergence, and is capable of generating high visual-quality images. Another contribution of BEGAN is providing an approximate measure of convergence. The earlier class of GANs lacks convergence measurement. Not until later, a new class of GANs exemplified by Wasserstein Generative Adversarial Network (WGAN)~\cite{WGAN} introduces a new loss metric, which correlates with the generator’s convergence. To our knowledge, BEGAN yields an alternative class of GANs that also has a loss correlated with convergence measurement.

    Apart from the class of energy-based GANs, Progressive Growing of Generative Adversarial Networks (PGGANs)~\cite{PGGAN} is another approach to generating high-quality images. By changing the training procedure without modifying the original GAN loss, PGGANs are able to increase training stability and to produce diverse yet high-resolution (up to $1024 \times 1024$ pixels) images.

    The $z^*$ approximation property of BEGAN-CS is similar to another class of bijective GANs, which constructs a bijection between the latent space $Z$ and the data space $X$. This class of models includes ALI~\cite{DumoulinBPLAMC16}, BiGAN~\cite{DonahueKD16}, VEEGAN~\cite{SrivastavaVRGS17} and \cite{reviewer_paper_A}. These four methods share a similar characteristic, requiring additional effort to optimize an extended network. VEEGAN introduces an extra reconstructor network $F_{\theta}$, which maps real data distribution $p(x)$ to a Gaussian. ALI/BiGAN both introduce an additional encoder network in the generator, and try to build up a bijection function. For \cite{reviewer_paper_A}, the loss term $L_s$ (Eq. (9) in \cite{reviewer_paper_A}) has a pre-requirement that the generator must include the real images in its latent space. They introduce an extra encoder network in generator to fulfill this requirement.

    In comparison, BEGAN-CS introduces a light-weight loss that utilizes the built-in mechanism of BEGAN without a need of extra networks. This makes the latent space inverting function jointly optimizable with the discriminator. Also, the constraint loss is a very strong indicator, detecting and protecting the model from mode collapsing. We also include further experimental comparisons with the class of bijective GANs in section \ref{subsection:comparison-with-bijective}.

\section{Methods}

    Mode collapse is a phenomenon that the generated images get stuck in or oscillate between a few modes. This phenomenon under BEGAN's setting has a unique characteristic. Since every sample shares the same encoder in the discriminator of BEGAN, the generated images that collapse at the same mode will share similar latent vectors as encoded by the encoder.
    
    By leveraging this property, we propose the \emph{latent-space constraint loss} ($\mathcal{L}_c$), or the \emph{constraint loss} for short. It constrains the norm of the difference between the latent vector $z$ and the internal state of encoder $Enc(G(z))$, where $Enc$ is the encoder within the discriminator. During the training process, the constraint loss is only optimized with respect to the discriminator. Although the mode-collapsing problem happens on the generator side, adding the constraint loss directly to the generator would expose too much information to the generator about how to exploit the discriminator, and thus turns out accelerating the occurrence of mode collapse. The constraint loss can also be viewed as a regularizer, which guides the function $Enc(G(\cdot))$ to be an identity function, and forces the encoder of the discriminator to retain the diversity and uniformity of randomly sampled $z \in Z$.

\begin{figure}[!t]
        \includegraphics[width=\linewidth]{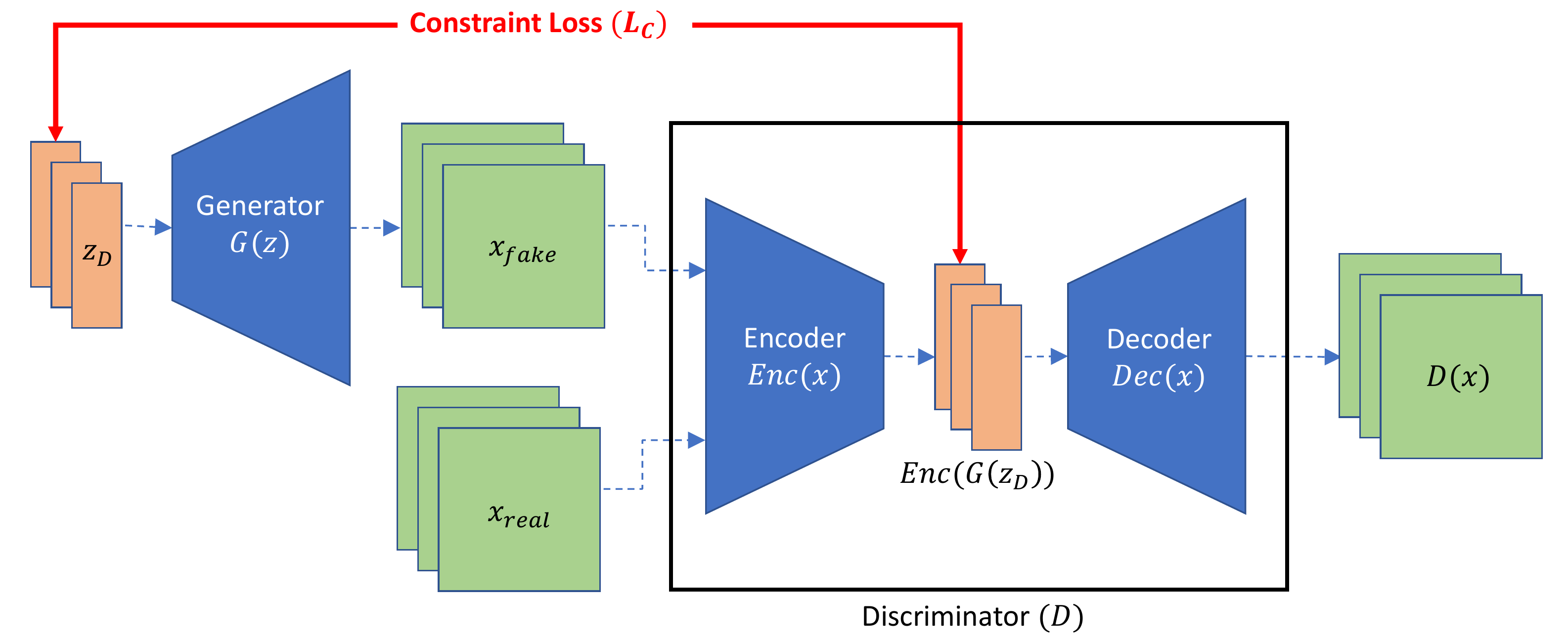}
        \caption{An overview of BEGAN-CS.}
        \label{fig:full-architecture}
    \end{figure}

    Fig.~\ref{fig:full-architecture} is an overview of the full-architecture of BEGAN-CS. The objective function of BEGAN-CS is mostly similar to BEGAN, except the additional constraint loss. The full objective of BEGAN-CS includes
    \begin{equation}
    \mathcal{L}_G = \mathcal{L}(G(z_G;\theta_G);\theta_D)\,, \quad  \mbox{for} \ \theta_G 
    \end{equation}
    \noindent and
    \begin{equation}
        \mathcal{L}_D = \mathcal{L}(x_\mathrm{real};\theta_D) - k_t \cdot \mathcal{L}(G(z_D;\theta_G);\theta_D) +  \alpha \cdot \mathcal{L}_c\,, \quad \mbox{for} \ \theta_D  \\
    \end{equation}
    \noindent with
    \begin{equation}
    \begin{cases}     
        \mathcal{L}_c = \lVert z_D - Enc(G(z_D)) \rVert \,, & \mbox{(the constraint loss)}\\
    
        k_{t+1} = k_{t} + \lambda (\gamma \mathcal{L}(x;\theta_D) - \mathcal{L}(G(z_G;\theta_G);\theta_D) )\,, & \mbox{for each epoch} \,.
        
    \end{cases}
    \end{equation}
    
    \noindent
    The total loss $\mathcal{L}_G$ of the generator and the total loss $\mathcal{L}_D$ of the discriminator are optimized to solve for the parameters $\theta_G$ and $\theta_D$, respectively. The function $\mathcal{L}(x;\theta_D) = \lVert x - D(x) \rVert$ associated with $\theta_D$ computes the norm of the difference between any given image $x$ and its reconstructed image $D(x)$ by the decoder of the discriminator. The latent vectors $z_D$ and $z_G$ are randomly sampled from $Z$. The variable $k_t \in [0,1]$ controls how much emphasis to put on $\mathcal{L}(G(z_D;\theta_G);\theta_D)$. The hyper-parameter $\gamma \in [0,1]$ balances between the real-image reconstruction loss $\mathcal{L}(x;\theta_D)$ and the generated-image discrimination loss $\mathcal{L}(G(z_G;\theta_G);\theta_D)$. The hyper-parameter $\alpha$ is a weighting factor for constraint loss. The constraint loss $\mathcal{L}_c$ is to enforce $Enc(G(\cdot))$ to be an identity function for $z_D$.

    \subsection{Latent Space Analysis}
    \label{subsection:PCA-analysis}
    
    For further illustrating the effectiveness of our method and analyzing the root cause of mode collapsing, we visualize the latent space through time with and without the constraint loss. We take PCA as our choice of dimensionality reduction method, and project the latent vectors onto two-dimensional space. Another common choice of dimensionality reduction for visualization is t-Distributed Stochastic Neighbor Embedding (t-SNE)~\cite{t-SNE}. For the latent space, we are more interested in the density and distribution of the points rather than the relative nearness between points or clusters. As a result, PCA is more suitable for our analysis.
    
\begin{figure}[!b]
        \centering
        \parbox[b]{\linewidth}{\Large
            \parbox[b]{.025\linewidth}{\rotatebox[origin=c]{90}{\small BEGAN}}%
            \subfloat{\begin{tabular}{c}
                \small Epoch 1 \\
                \includegraphics[width=0.185\linewidth,height=0.18\linewidth]{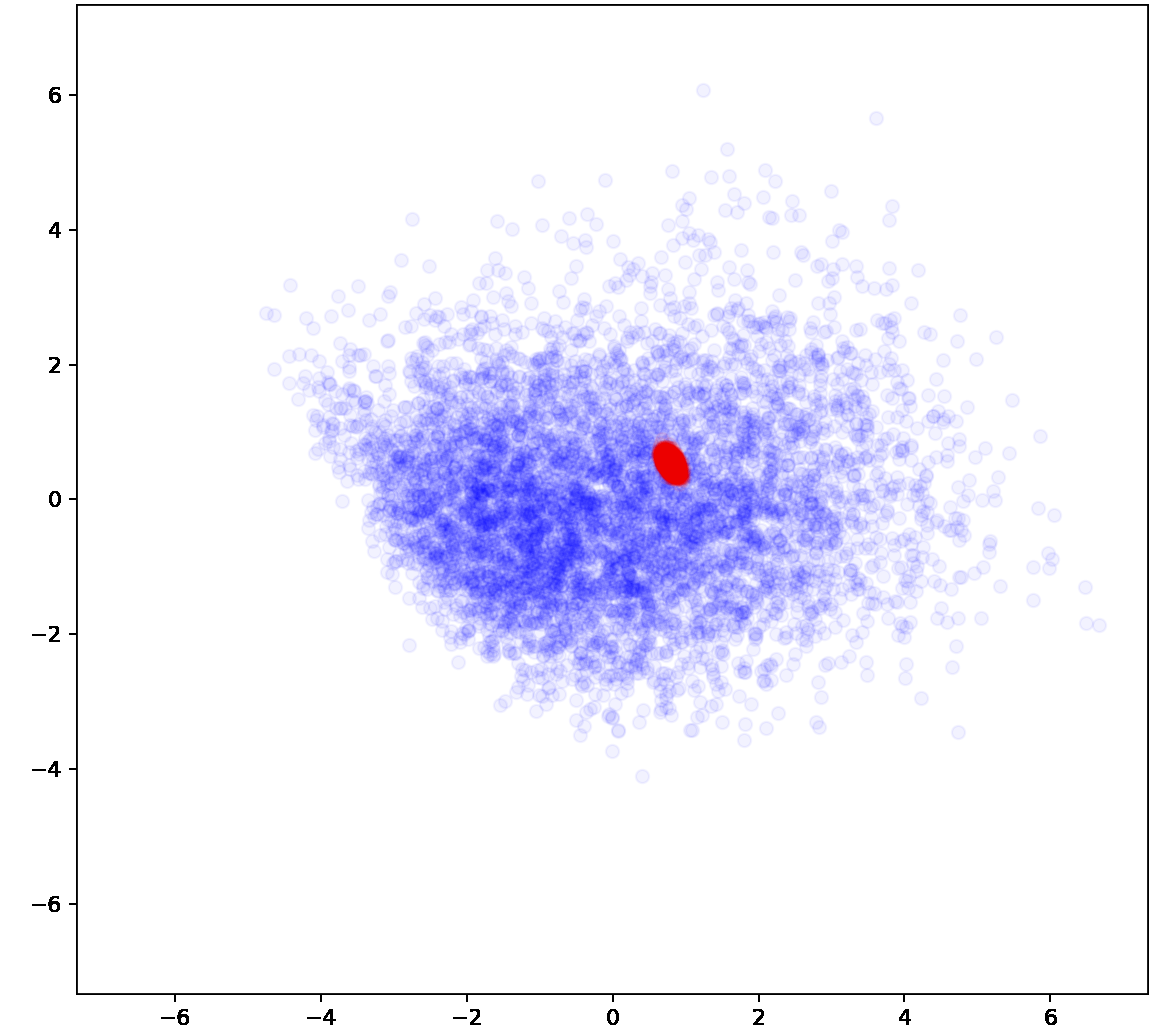} \\
                [-0.8em] \tiny $\text{Var}(real) = 4.21$ \\
                [-0.8em] \tiny $\text{Var}(gen) = 0.60$
            \end{tabular}}\hfill%
            \subfloat{\begin{tabular}{c}
                \small Epoch 11 \\
                \includegraphics[width=0.185\linewidth,height=0.18\linewidth]{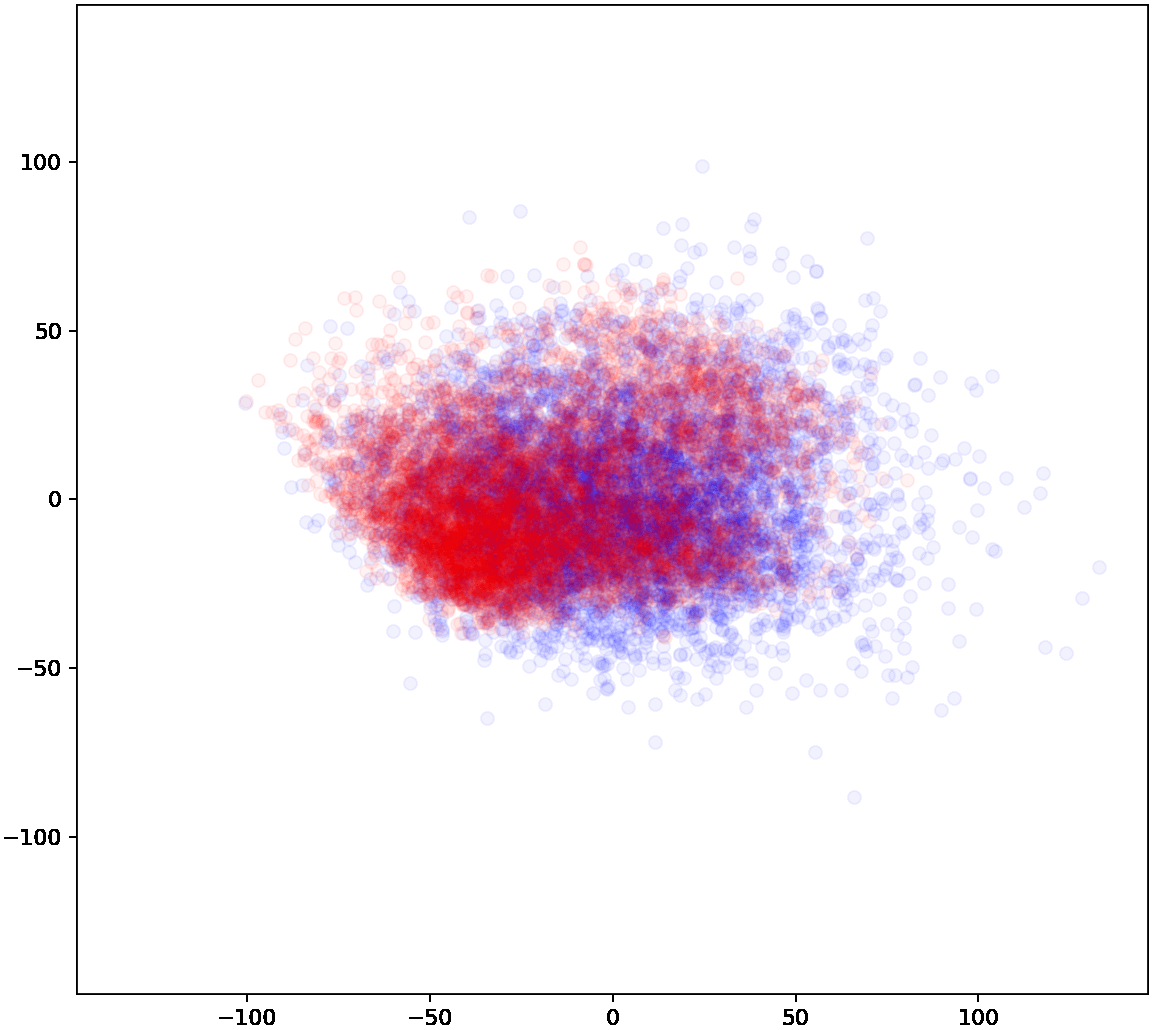} \\
                [-0.8em] \tiny $\text{Var}(real) = 85.05$ \\
                [-0.8em] \tiny $\text{Var}(gen) = 68.64$
            \end{tabular}}\hfill%
            \subfloat{\begin{tabular}{c}
                \small Epoch 21 \\
                \includegraphics[width=0.185\linewidth,height=0.18\linewidth]{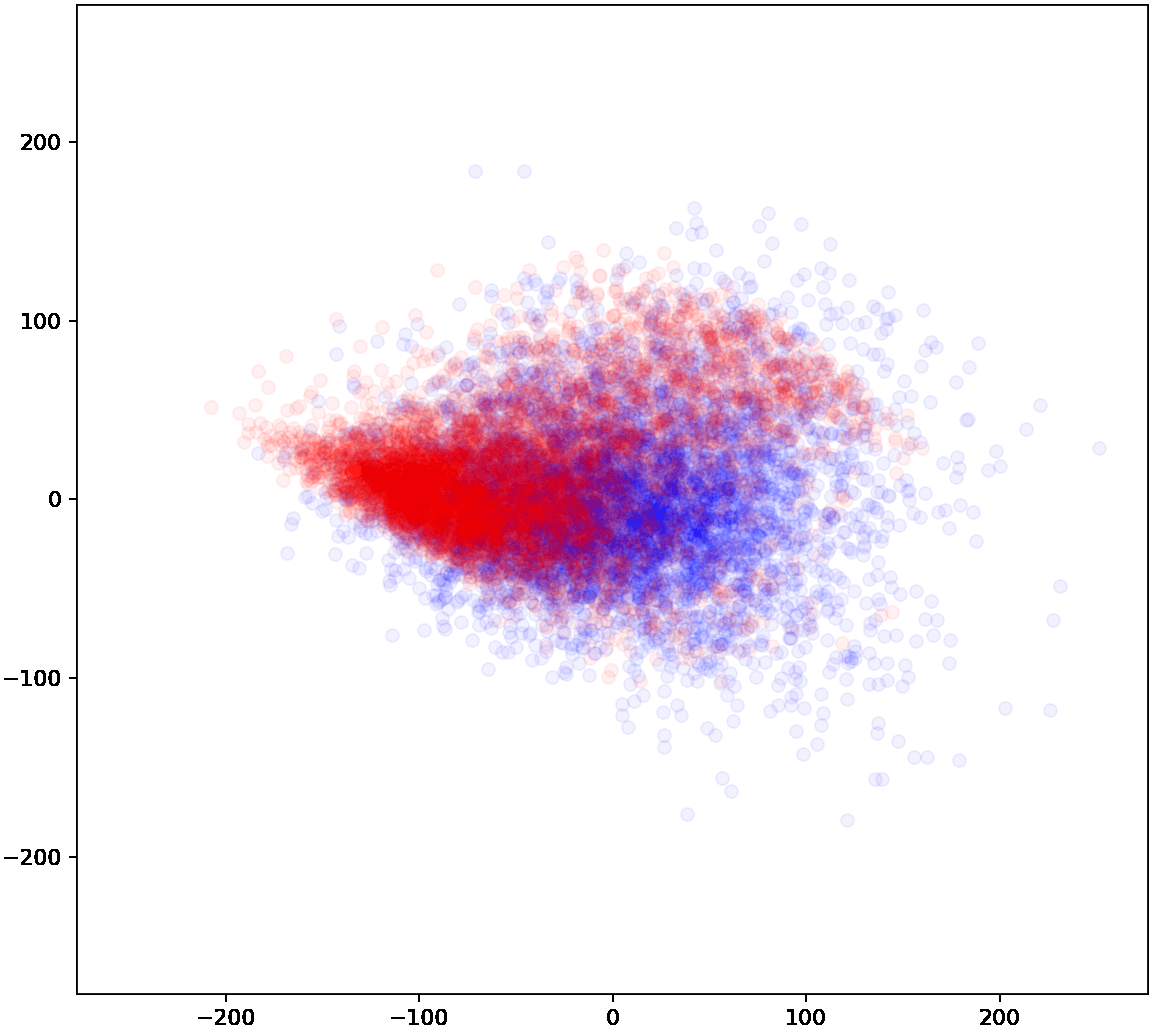} \\
                [-0.8em] \tiny $\text{Var}(real) = 153.42$ \\
                [-0.8em] \tiny $\text{Var}(gen) = 123.86$
            \end{tabular}}\hfill%
            \subfloat{\begin{tabular}{c}
                \small Epoch 31 \\
                \includegraphics[width=0.185\linewidth,height=0.18\linewidth]{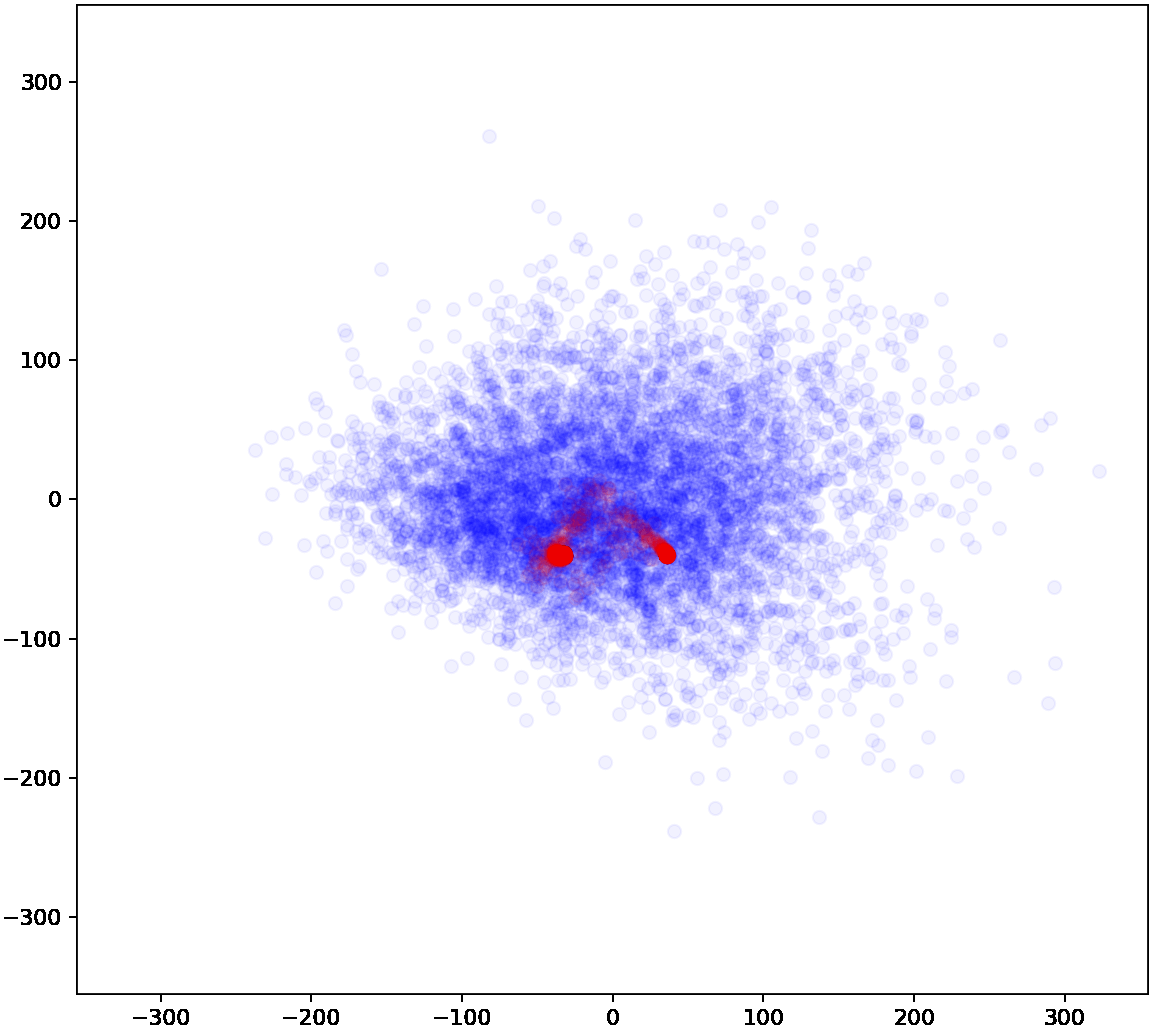} \\
                [-0.8em] \tiny $\text{Var}(real) = 214.25$ \\
                [-0.8em] \tiny $\text{Var}(gen) = 134.32$
            \end{tabular}}\hfill%
            \subfloat{\begin{tabular}{c}
                \small Epoch 41 \\
                \includegraphics[width=0.185\linewidth,height=0.18\linewidth]{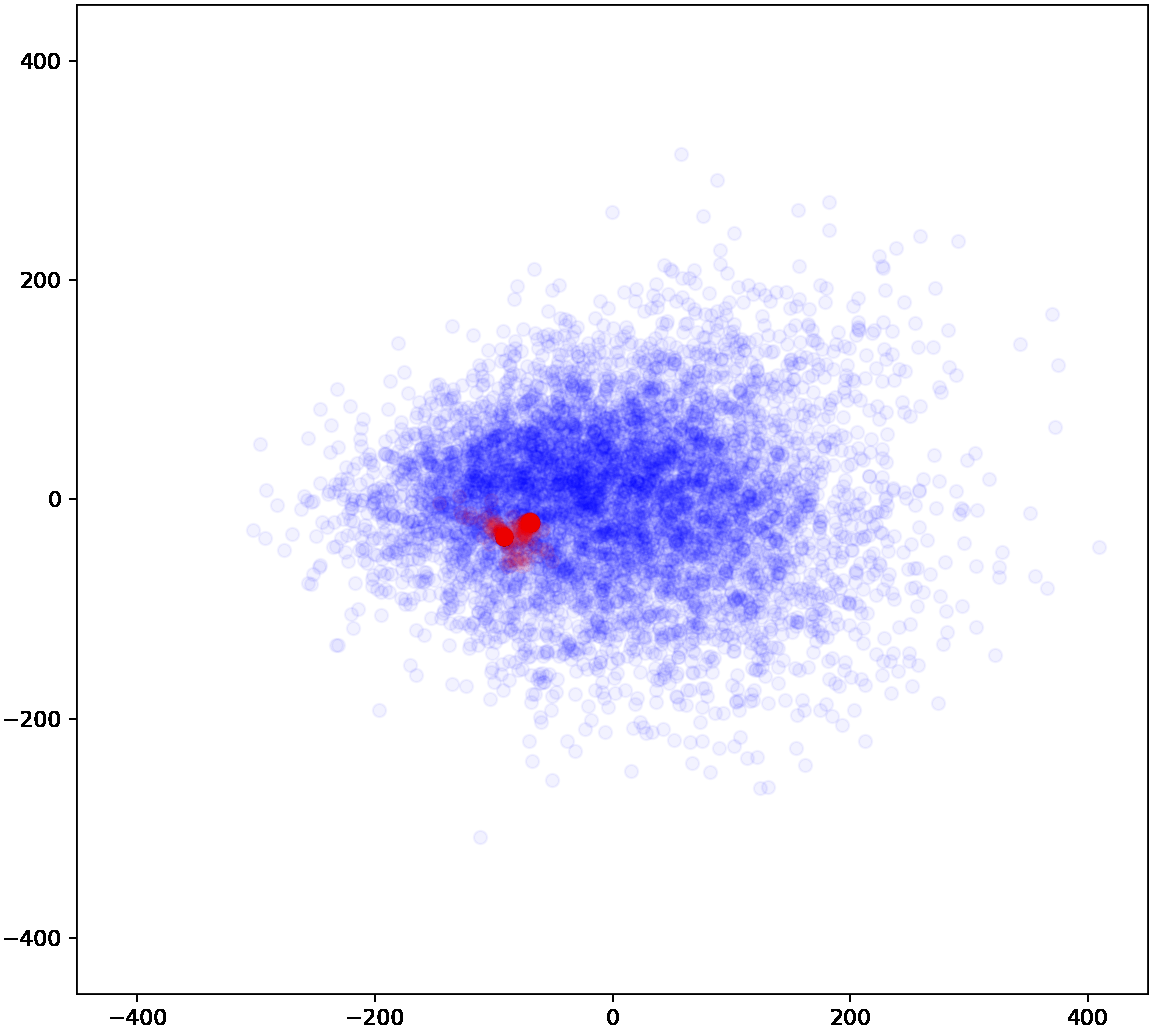} \\
                [-0.8em] \tiny $\text{Var}(real) = 268.11$ \\
                [-0.8em] \tiny $\text{Var}(gen) = 159.71$
            \end{tabular}}
            
            \parbox[b]{.025\linewidth}{\rotatebox[origin=c]{90}{\small BEGAN-CS}}%
            \subfloat{\begin{tabular}{c}
                \small Epoch 1 \\
                \includegraphics[width=0.185\linewidth,height=0.18\linewidth]{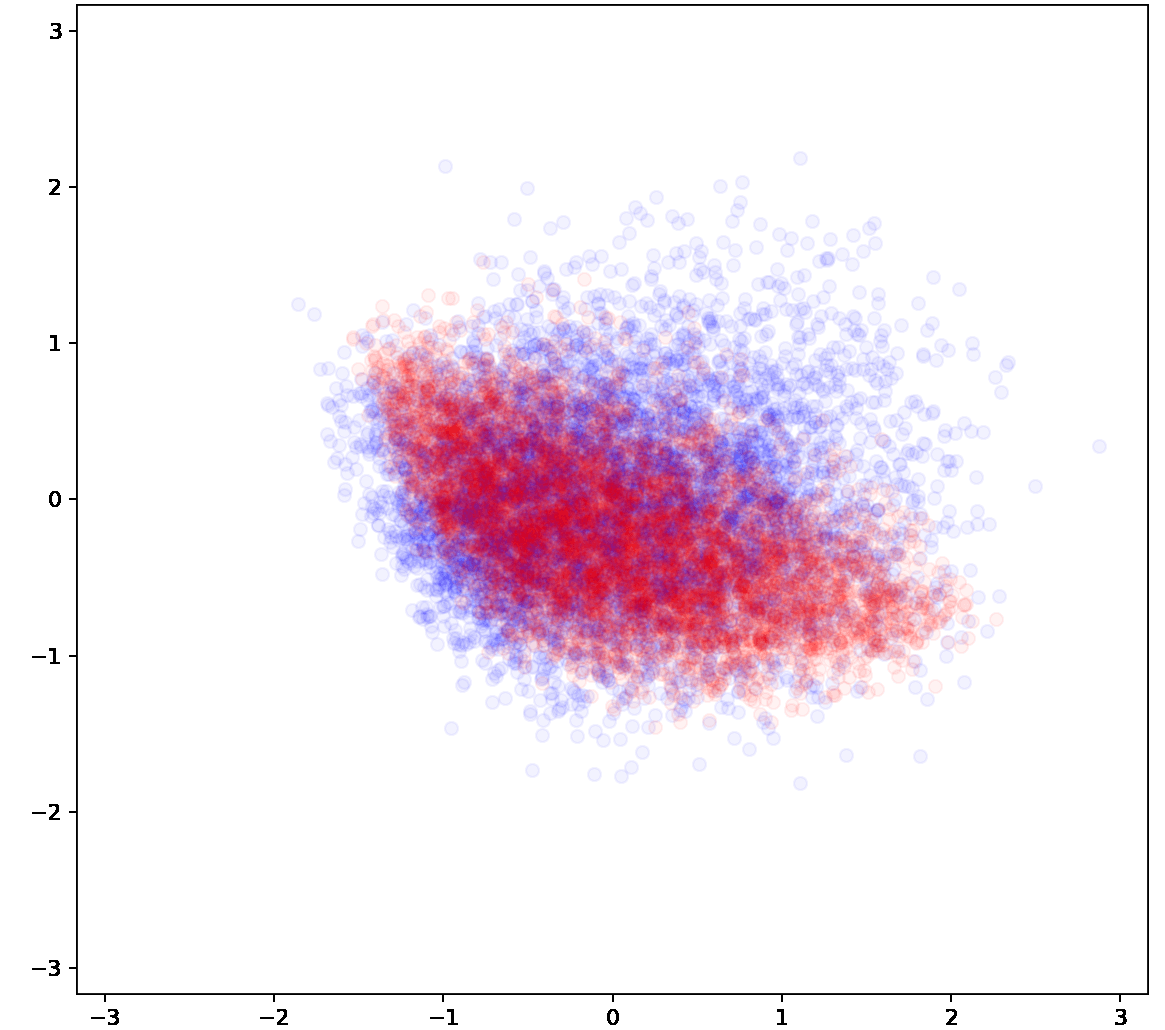} \\
                [-0.8em] \tiny $\text{Var}(real) = 2.0$ \\
                [-0.8em] \tiny $\text{Var}(gen) = 1.40$
            \end{tabular}}\hfill%
            \subfloat{\begin{tabular}{c}
                \small Epoch 11 \\
                \includegraphics[width=0.185\linewidth,height=0.18\linewidth]{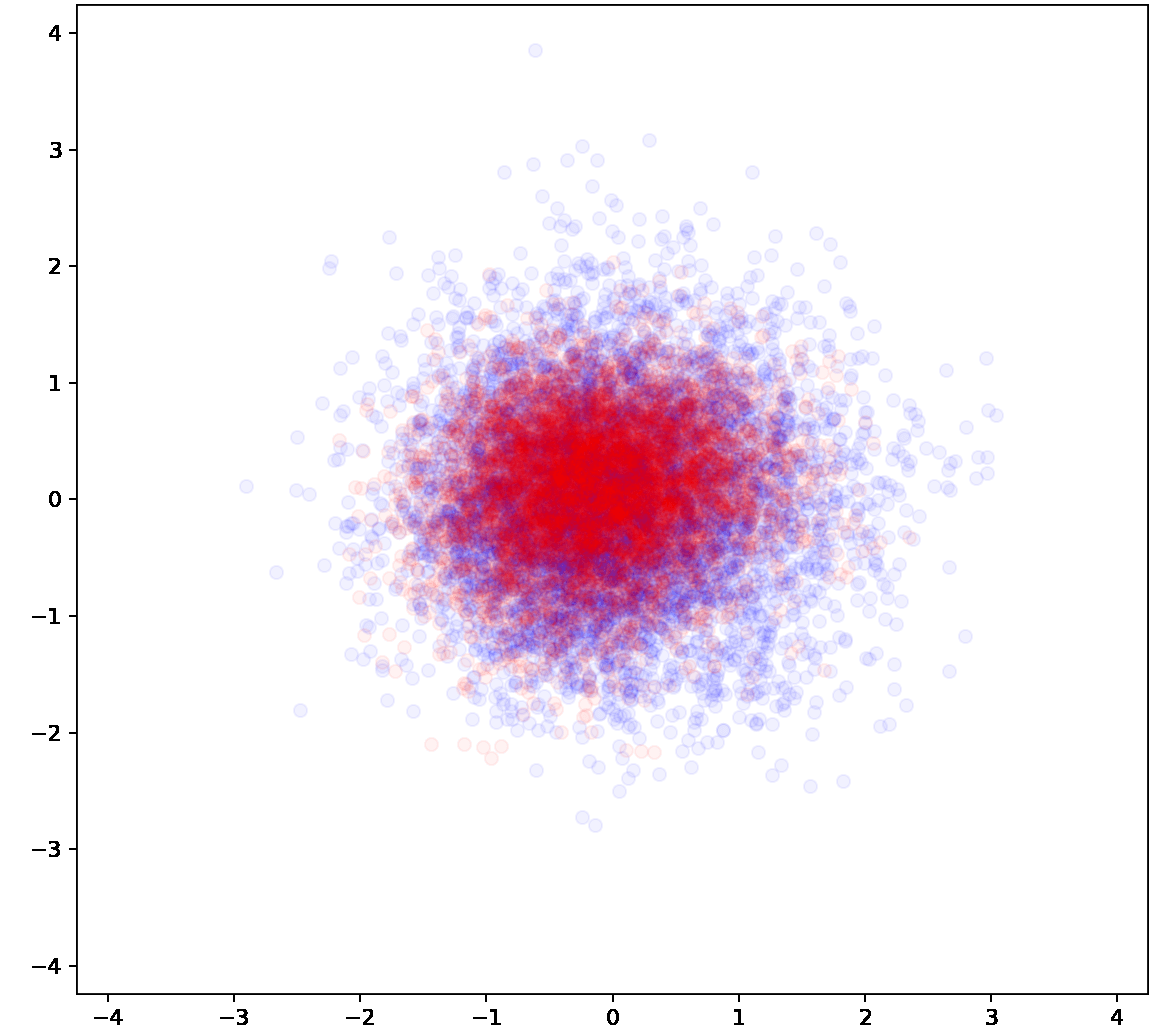} \\
                [-0.8em] \tiny $\text{Var}(real) = 3.74$ \\
                [-0.8em] \tiny $\text{Var}(gen) = 3.17$
            \end{tabular}}\hfill%
            \subfloat{\begin{tabular}{c}
                \small Epoch 21 \\
                \includegraphics[width=0.185\linewidth,height=0.18\linewidth]{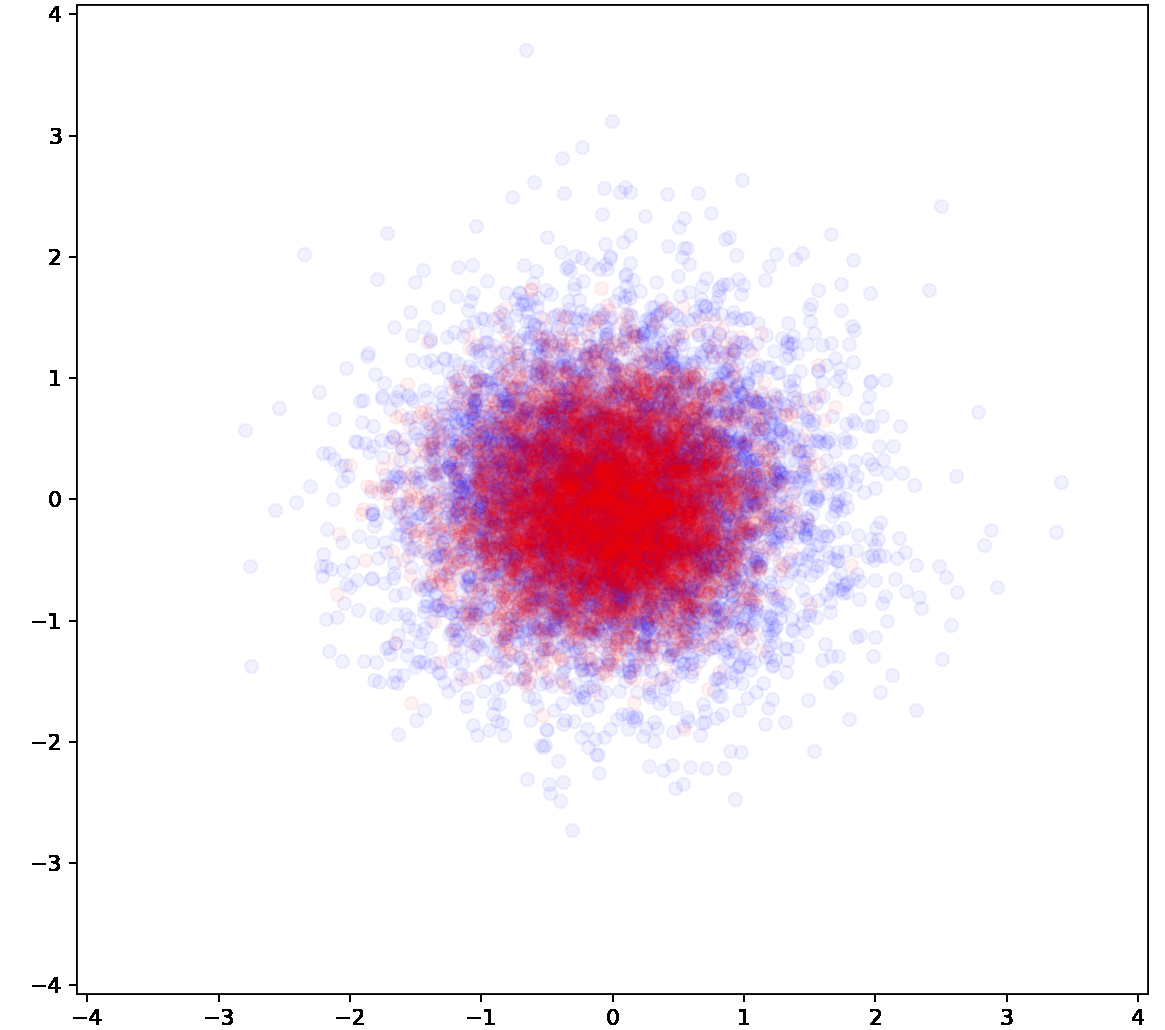} \\
                [-0.8em] \tiny $\text{Var}(real) = 3.62$ \\
                [-0.8em] \tiny $\text{Var}(gen) = 2.90$
            \end{tabular}}\hfill%
            \subfloat{\begin{tabular}{c}
                \small Epoch 31 \\
                \includegraphics[width=0.185\linewidth,height=0.18\linewidth]{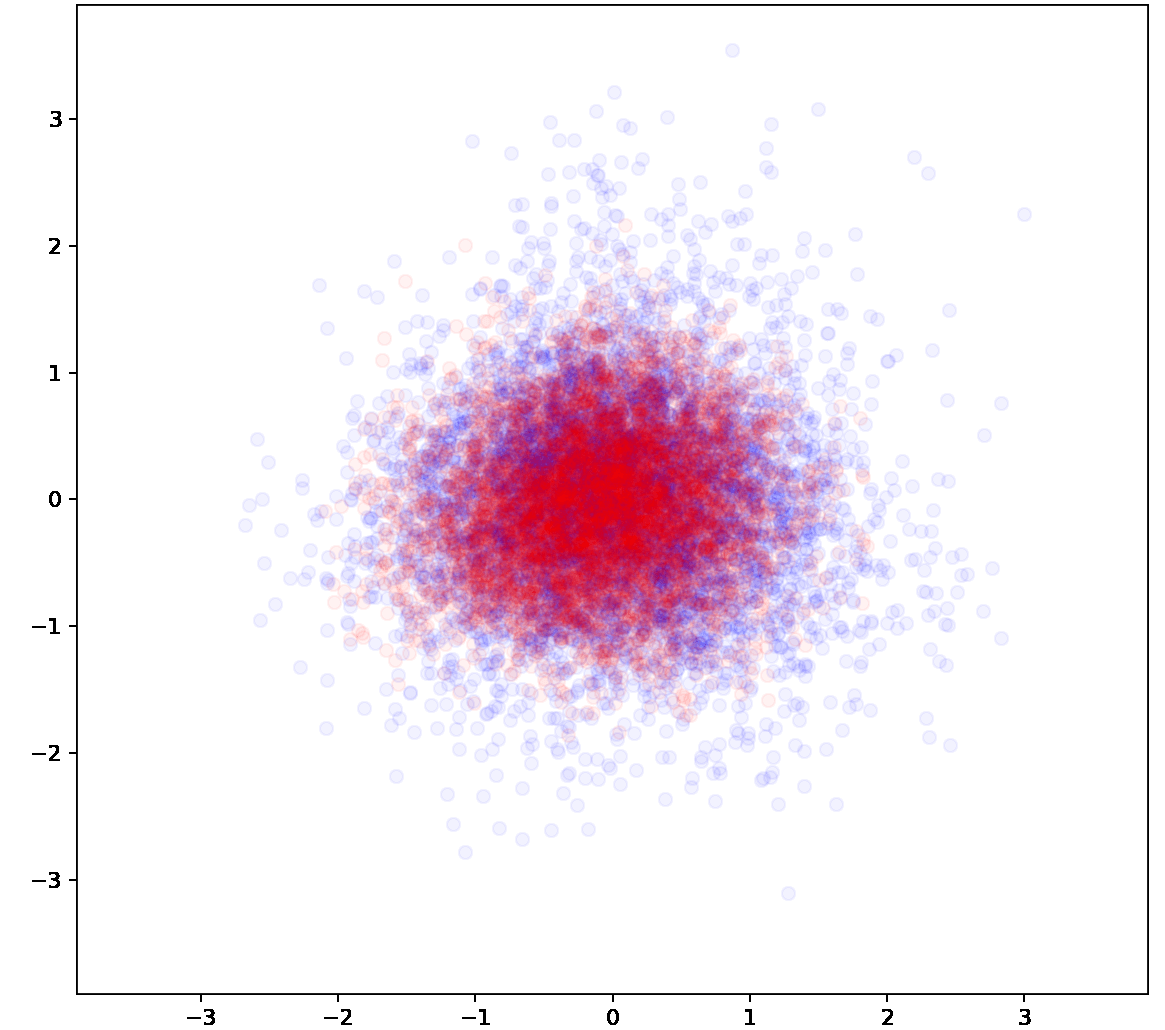} \\
                [-0.8em] \tiny $\text{Var}(real) = 3.56$ \\
                [-0.8em] \tiny $\text{Var}(gen) = 2.88$
            \end{tabular}}\hfill%
            \subfloat{\begin{tabular}{c}
                \small Epoch 101 \\
                \includegraphics[width=0.185\linewidth,height=0.18\linewidth]{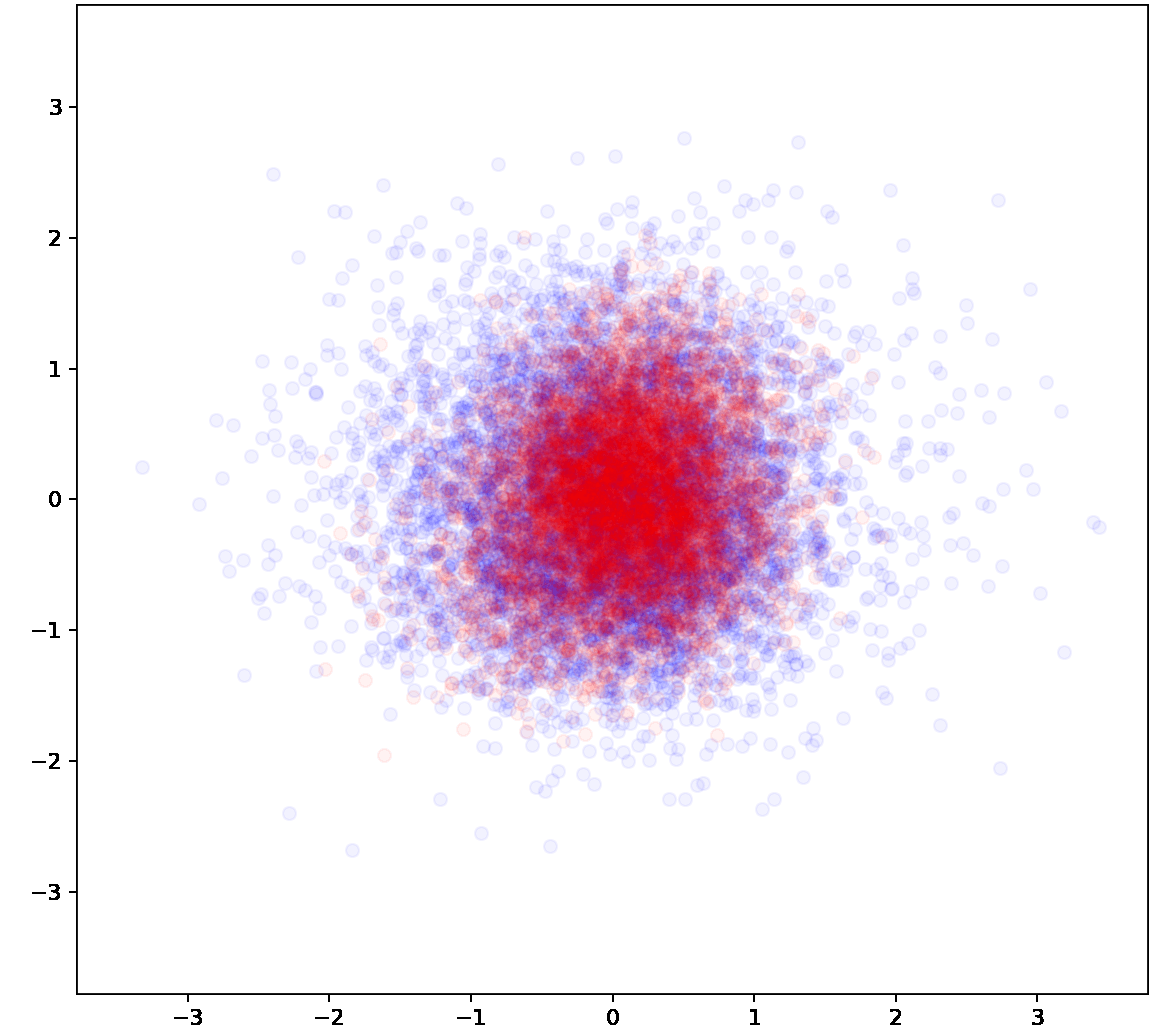} \\
                [-0.8em] \tiny $\text{Var}(real) = 3.10$ \\
                [-0.8em] \tiny $\text{Var}(gen) = 2.50$
            \end{tabular}}
            
        }
        
        \caption{We visualize the distributions of latent vectors of BEGAN and BEGAN-CS over epochs. Both BEGAN and BEGAN-CS are trained on CelebA dataset under $64 \times 64$ resolution and batch size 64. Each graph consists of 6{,}400 random real images' latent vectors, \ie $Enc(x)$, and 6{,}400 generated images' latent vectors, \ie $Enc(G(z))$. The upper five graphs are generated by BEGAN, while the bottom five graphs are produced by BEGAN-CS. PCA is performed separately at each epoch based on the latent vectors of the real images. Each blue point represents a latent vector of a real image after applying PCA, and the red points correspond to the latent vectors of the generated images. The text under each graph lists the variance of real images' latent vectors (Var(real)) and the variance of generated images' latent vectors (Var(gen)). During the training of BEGAN, the variances of the distributions of latent vectors keep growing. Note that most of the graphs are created with a fixed interval of 10 epochs, except the bottom-right graph directly skips to the 101st epoch to highlight the effectiveness of BEGAN-CS. BEGAN has already collapsed before the 41st epoch.}\label{fig:PCA-analysis}
    \end{figure}

    Fig.~\ref{fig:PCA-analysis} shows a preliminary analysis of BEGAN and BEGAN-CS. We train both models on the CelebA dataset~\cite{CelebA}. The 64-dimensional latent vectors of generated images ($Enc(G(z))$) and real images ($Enc(x)$) are projected onto two-dimensional space via PCA. 
    
    In this experiment, BEGAN gets into mode collapse at epoch 23. In addition to the obvious change in the shape of distribution after BEGAN mode-collapsing, our empirical analysis also shows two strong patterns. First, in comparison with BEGAN, the latent-vector distribution (in red) of images generated by BEGAN-CS can better fit the real images' latent-vector distribution (in blue). The latent vectors of BEGAN-CS scatter more uniformly across all epochs. 
    
    Second, for BEGAN without adding the constraint loss, both the variance of real images' latent vectors (Var(real)) and the variance of generated images' latent vectors (Var(gen)) grow rapidly as the number of epochs increases. Our hypothesis is that the latent spaces of real images and generated images  both expand too rapidly and non-uniformly. Since the number of training data is fixed, as the latent space of real images expands, the density of real images decreases. In the end, the generator of BEGAN reaches a low-density area in the latent space where there is only a few latent vectors of real images nearby. The generator of BEGAN then gets stuck in that area. In contrast, BEGAN-CS has the latent-space constraint as a regularizer, which restricts the latent spaces of real images and generated images expand incautiously. In other words, the constraint loss limits the distribution of $Enc(G(z))$ to be similar to uniform distribution.

    \subsection{Obtaining Optimal $z^*$ in One-Shot}
    \label{subsection:obtain-z-in-one-shot}
    
    Given an image $x^*$, finding an optimal latent vector $z^*$ such that $\lVert G(z^*)- x^* \rVert < \epsilon$ for some small $\epsilon$ is a challenging problem for GANs. Traditionally, $z^*$ can be obtained by back-propagation for solving the optimization $\displaystyle \min_{z^*} (\lVert G(z^*) - x^* \rVert)$. We name this optimization process as $z^*$-search. However, $z^*$-search is  time-consuming and needs to run for each inference individually, and thus is impractical for real-world applications.
    
    In the case of BEGAN-CS, the constraint loss works as a regularizer, guiding the composite function $Enc(G(z)) \simeq z$ to be similar to an identity function.
    Consider the definition of $z^*$, where $G(z^*) = x^*$. We know that $Enc(G(z^*))$ should be close to $z^*$ due to the identity property. This implies that we may take $x^*$ and obtain $Enc(x^*)$ as an approximation to $z^*$ after a single pass through the encoder $Enc(x^*)$.
    
    \subsection{Disentangled Representation Learning and Application}
    \label{subsection:method-disentangled-representation-and-application}
    
    We find that BEGAN is able to learn strong and high-quality disentangled representations in an unsupervised setting. The direction of any vector within latent space $Z$ has a universally meaningful semantic, such as mixture of gender, age, smile and hair-style. These learned representations can be combined with vector arithmetic operations to generate images with multiple designated representations.
    
\begin{figure}[!t]
        \centering
        \begin{tabular}{L{0.6\linewidth}R{0.25\linewidth}R{0.05\linewidth}}
            \large{$G ( \, Enc ( $ \raisebox{-.42\height}{\includegraphics[width=1.2cm]{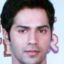}} $ ) \, ) $} & \large{ $\Longrightarrow$ \, \raisebox{-.42\height}{\includegraphics[width=1.2cm]{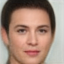}}} & (a) \\
            \\ [-0.5em]
            \large{$G ( \, Enc ( $ \raisebox{-.42\height}{\includegraphics[width=1.2cm]{imgs/combine_example/real_image.png}} $ ) \, + \, style A \, ) $} & \large{ $\Longrightarrow$ \, \raisebox{-.42\height}{\includegraphics[width=1.2cm]{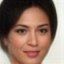}}} & (b)\\
            \\ [-0.5em]
            \large{$G ( \, Enc ( $ \raisebox{-.42\height}{\includegraphics[width=1.2cm]{imgs/combine_example/real_image.png}} $ ) \, + \, style B \, ) $} & \large{ $\Longrightarrow$ \, \raisebox{-.42\height}{\includegraphics[width=1.2cm]{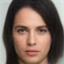}}} & (c)\\
            \\ [-0.5em]
            \large{$G ( \, Enc ( $ \raisebox{-.42\height}{\includegraphics[width=1.2cm]{imgs/combine_example/real_image.png}} $ ) \, + \, style A \, + \, style B \, )$} & \large{ $\Longrightarrow$ \, \raisebox{-.42\height}{\includegraphics[width=1.2cm]{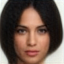}}} & (d)
        \end{tabular}
        
        \caption{An example of disentangled representations. The ``$style A$'' and ``$style B$'' are two learned disentangled representations. Note that these representations are universal and can be applied to any latent vector $z$ for generating images $G(z + style)$ with designated attributes. (a) Approximate $z^*$ by $G(Enc(x^*))$ in one-shot. (b) \& (c) The learned disentangled representations can be combined with $G(Enc(x^*))$. (d) Vector arithmetic with multiple disentangled representations. In this case, the generated image has both hair-styles shown in $style A$ and $style B$.}
        \label{fig:combine-example}
        
    \end{figure}

    However, these disentangled representations are only effective for latent vectors, which is a strong restriction that forbids many GAN models to use the disentangled representation for practical applications, since obtaining the latent vectors via $z^*$-search is computation-demanding. In the meanwhile, as we have shown in section \ref{subsection:obtain-z-in-one-shot}, BEGAN-CS is able to produce the approximation of $z^*$ on-the-fly. By adding multiple selected representation vectors to the approximated $z^*$ with respect to any given real image $x^*$, we can generate images that are visually similar to $x^*$ and comprise the selected representations at the same time. We demonstrate this idea with a real example produced by BEGAN-CS in Fig~\ref{fig:combine-example}. In this example, the generated image of Fig~\ref{fig:combine-example}d acquires both hair-styles shown Fig~\ref{fig:combine-example}b \& Fig~\ref{fig:combine-example}c. For BEGAN, which lacks the ability of estimating $z^*$ directly, the same effect may be forcibly achieved through time-consuming $z^*$-search to obtain suitable $z^*$. Unfortunately, $z^*$-search causes the major bottleneck at inference time and is therefore hard to use in real-world scenarios.
    
    Similar applications can also be achieved using Variational Auto-Encoder (VAE) based models~\cite{VAE,beta-VAE,VAE-GAN} or other task-specific GAN models, such as InfoGAN~\cite{info-GANs}. However, the images generated by VAE-based models tend to be blurry, while InfoGAN cannot generate high-quality results as BEGAN does. In comparison, our results are more promising in terms of stability and quality.

\section{Experiments}
\label{section:experiments}

    We train BEGAN-CS using the CelebA dataset for all the experiments presented in this paper. BEGAN-CS does not adopt the learning rate decay technique described in BEGAN's original paper, since the training process of BEGAN-CS is already very stable. The hyper-parameter $\alpha$ that controls the importance of the constraint loss is set to $0.1$ as the default value. 
    We use L2-norm in $$\mathcal{L}(x;\theta_D) = \lVert x - D(x) \rVert$$ throughout the experiments, while in practice, L1-norm can also be used.
    For any hyper-parameter that is not mentioned, we choose the same value as in  BEGAN's original setting. 
    
    \subsection{Effectiveness of the Constraint Loss}
    
\begin{figure}[!t]
        
        \parbox[b]{\linewidth}{\Large
            \parbox[b]{.03\linewidth}{\rotatebox[origin=c]{90}{\small BEGAN}}%
            \subfloat{\begin{tabular}{c}
            \small (A) \\
            \includegraphics[width=0.285\linewidth]{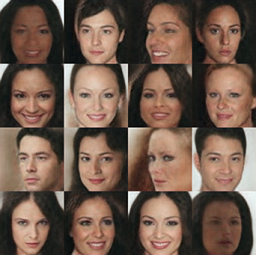}
            \end{tabular}}\hfill%
            \subfloat{\begin{tabular}{c}
            \small (B) \\
            \includegraphics[width=0.285\linewidth]{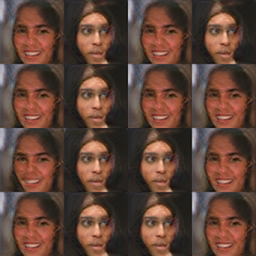}
            \end{tabular}}\hfill%
            \subfloat{\begin{tabular}{c}
            \small (C) \\
            \includegraphics[width=0.285\linewidth]{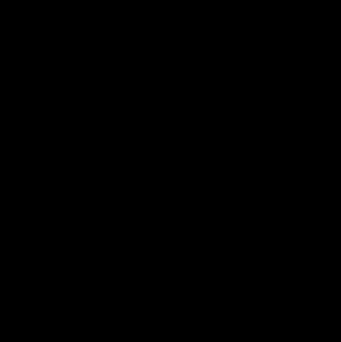}
            \end{tabular}}\hfill%
        }
        
                    
        \centering
        \parbox[b]{\linewidth}{\Large
            \parbox[b]{.03\linewidth}{\rotatebox[origin=c]{90}{\small BEGAN-CS}}%
            \subfloat{\begin{tabular}{c}
            \includegraphics[width=0.285\linewidth]{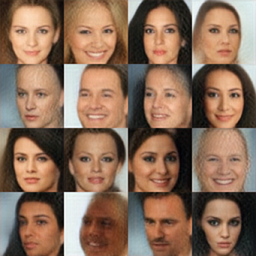}
            \end{tabular}}\hfill%
            \subfloat{\begin{tabular}{c}
            \includegraphics[width=0.285\linewidth]{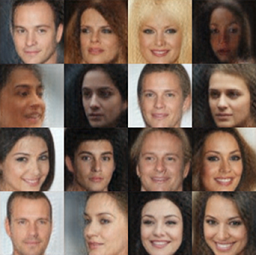}
            \end{tabular}}\hfill%
            \subfloat{\begin{tabular}{c}
            \includegraphics[width=0.285\linewidth]{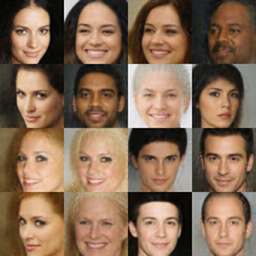}
            \end{tabular}}\hfill%
        }
        
        \subfloat{
            \includegraphics[height=0.32\linewidth]{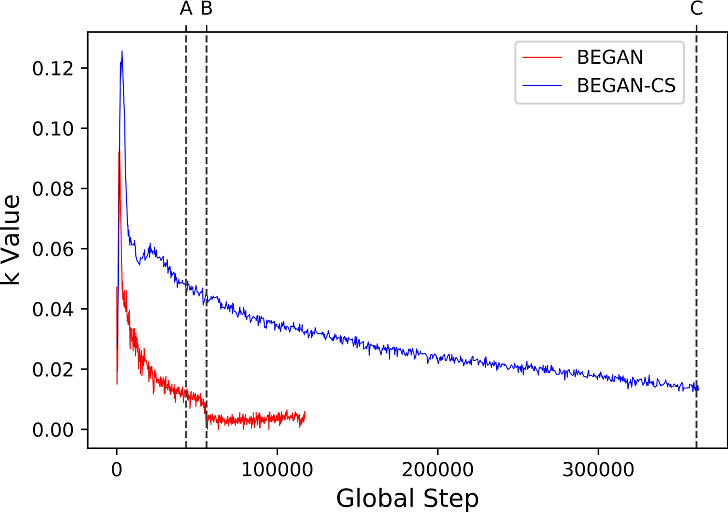}}%
        \hfill%
        \subfloat{
            \includegraphics[height=0.32\linewidth]{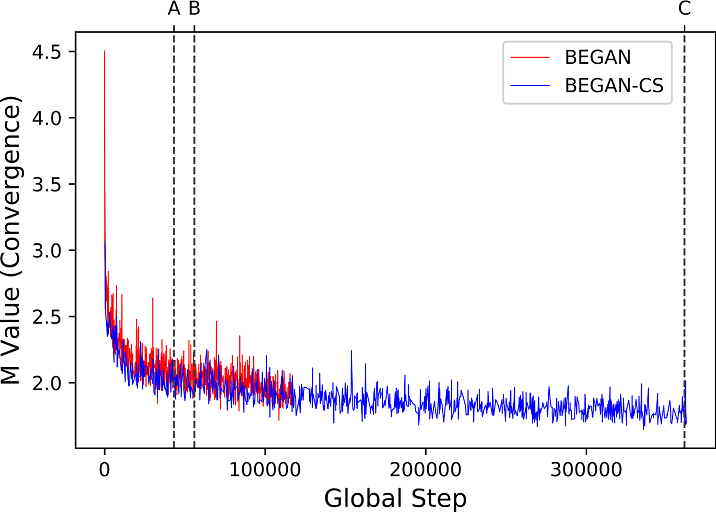}}
        
        \caption{We validate the effectiveness of the constraint loss by showing the generated images at specific epochs during the training of BEGAN and BEGAN-CS on the CelebA dataset. The image resolution is $64 \times 64$ and the batch size is 64. Note that BEGAN fails to reach epoch {\sf C} since it already collapses at epoch {\sf B}. In contrast, BEGAN-CS survives after epoch {\sf C}. Furthermore, BEGAN-CS maintains a very good performance in generating high-quality images.}
        \label{fig:experiment-manual-training-compare}
    \end{figure}

    In Fig.~\ref{fig:experiment-manual-training-compare}, we validate the effectiveness of the constraint loss. We show the generated images at specific epochs during the training of BEGAN and BEGAN-CS on the CelebA dataset. The image resolution is $64 \times 64$ and the batch size is 64.  BEGAN-CS can continuously be trained up to 100 epochs without any evidence of mode collapsing, loss of diversity, or reduction in quality. In contrast, BEGAN encounters mode collapse at the 25th epoch (\ie, the time-step {\sf B} in Fig.~\ref{fig:experiment-manual-training-compare}). In addition to the advantage of preventing from mode collapse, the proposed BEGAN-CS model also maintains a very good performance in generating high-quality images.

    \subsection{Observing the Sudden Mode Collapsing}
    \label{subsection:SuddenModeCollapse}
    
    An interesting finding during our experiments is the timing of mode collapsing.
    As is mentioned in \cite{BEGAN}, the global measure of convergence can be used by BEGAN to determine whether the network has reached the final state or if the model has collapsed. However, in practice we are not able to observe significant evidence of mode collapsing directly from the value of the convergence measure. Instead, the evidence of mode collapse are more often to be observed from the $k$ value. The $k$ value in BEGAN controls how much attention is paid on $\mathcal{L}(G(z))$. According to our observation, every time the $k$ value suddenly drops, BEGAN is going to collapse shortly.

    \subsection{Better Convergence on Small Datasets}
    \label{subsection:better-convergence-on-small-dataset}
    
\begin{figure}[!t]
        
        \parbox[b]{\linewidth}{\Large
            \parbox[b]{.03\linewidth}{\rotatebox[origin=c]{90}{\small BEGAN}}%
            \subfloat{\begin{tabular}{c}
            \small (A)\\
            \includegraphics[width=0.285\linewidth]{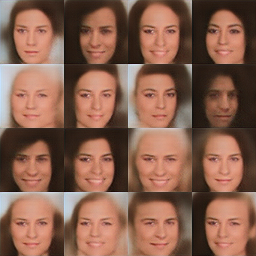}
            \end{tabular}}\hfill%
            \subfloat{\begin{tabular}{c}
            \small (B)\\
            \includegraphics[width=0.285\linewidth]{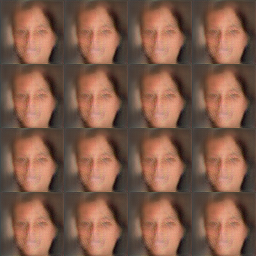}
            \end{tabular}}\hfill%
            \subfloat{\begin{tabular}{c}
            \small (C)\\
            \includegraphics[width=0.285\linewidth]{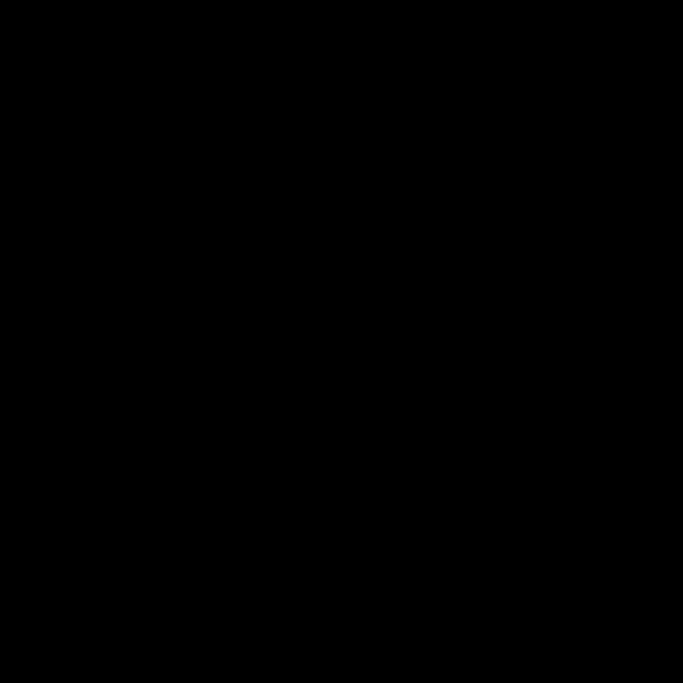}
            \end{tabular}}
        } 
        
        \centering
        \parbox[b]{\linewidth}{\Large
            \parbox[b]{.03\linewidth}{\rotatebox[origin=c]{90}{\small BEGAN-CS}}%
            \subfloat{\begin{tabular}{c}
            \includegraphics[width=0.285\linewidth]{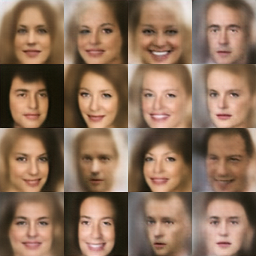}
            \end{tabular}}\hfill%
            \subfloat{\begin{tabular}{c}
            \includegraphics[width=0.285\linewidth]{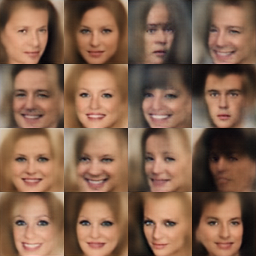}
            \end{tabular}}\hfill%
            \subfloat{\begin{tabular}{c}
            \includegraphics[width=0.285\linewidth]{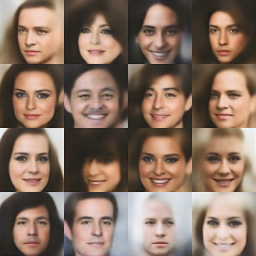}
            \end{tabular}}
        }
        
        \subfloat{
            \includegraphics[height=0.32\linewidth]{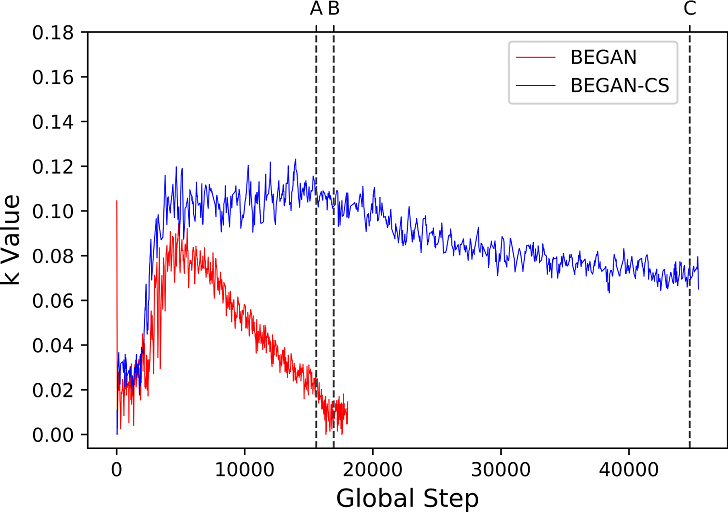}}%
        \hfill%
        \subfloat{
            \includegraphics[height=0.32\linewidth]{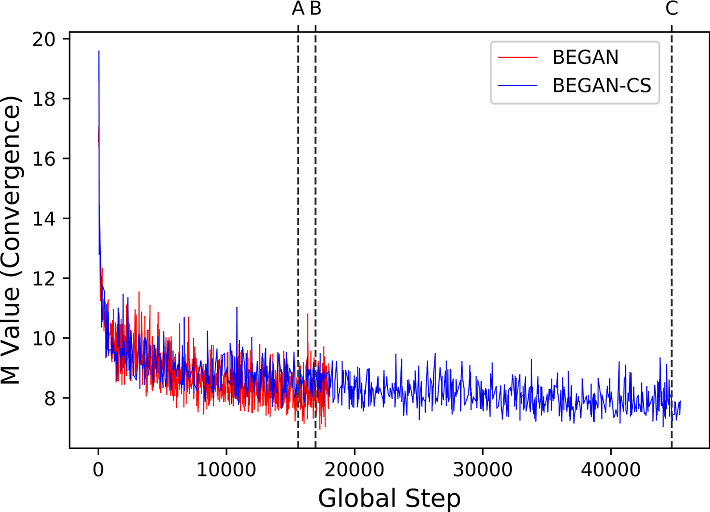}}
        
        \caption{Better convergence of BEGAN-CS on small datasets. We show the generated images at selected epochs during training BEGAN and BEGAN-CS on a $1/10$ sized subset of CelebA. Training images are of $128 \times 128$ resolution and the batch size is 24. BEGAN-CS is stable and converges to a particularly better state than BEGAN. The best state of BEGAN is at time-step {\sf A} with degraded quality, while BEGAN-CS can generate higher-quality results at time-step {\sf C}.}
        \label{fig:experiment-small-data-training-compare}
    \end{figure}

    The dataset size is also an important factor for the timing of mode collapse. Under a setting of reducing the training dataset CelebA to $1/10$ of its original size, BEGAN collapses earlier than training on full dataset. The early occurrence of mode collapse keeps BEGAN from converging to an optimal state. The time-step {\sf A} in Fig.~\ref{fig:experiment-small-data-training-compare} is the best state that BEGAN can achieve during its training on the down-sized CelebA dataset. On the other hand, BEGAN-CS has a more stable training process. In Fig.~\ref{fig:experiment-small-data-training-compare}, BEGAN-CS can continuously optimize on the $1/10$ down-sized CelebA dataset without encountering  mode collapse, and eventually converges to a better state than BEGAN.
    
    \subsection{FID Score Curve Comparison}
    
    \newlength{\oldintextsep}
    \setlength{\oldintextsep}{\intextsep}
    \setlength\intextsep{0pt}
    
    \begin{wrapfigure}[13]{r}{0.65\linewidth}
        \captionsetup{width=0.8\linewidth}
        \centering
        \begin{tabular}{C{0.5\linewidth}C{0.5\linewidth}}
             \includegraphics[width=0.95\linewidth]{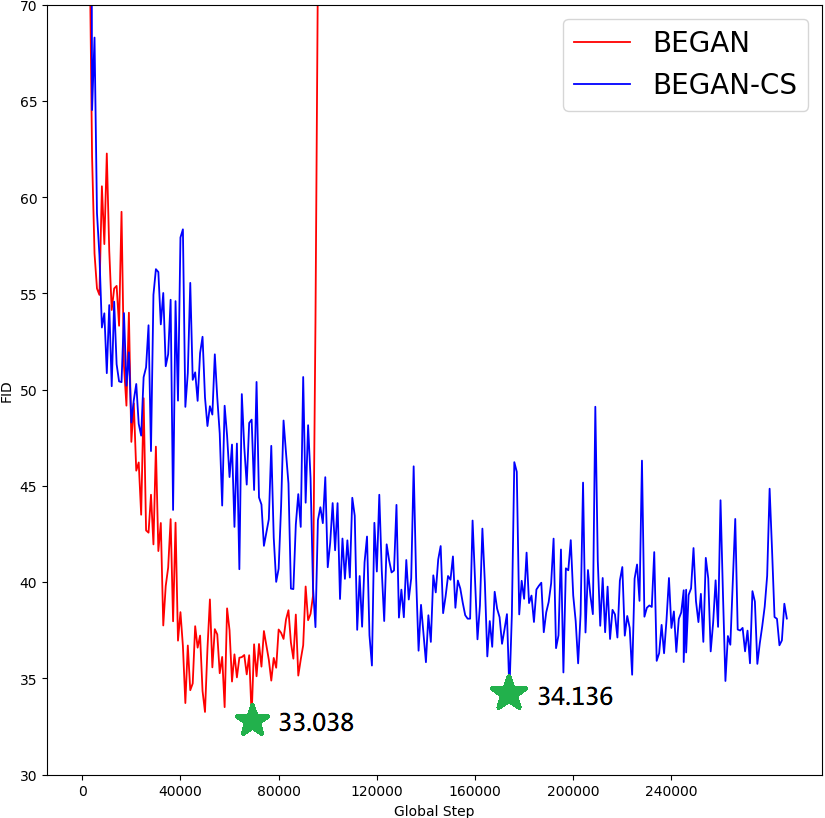} &
             \includegraphics[width=0.95\linewidth]{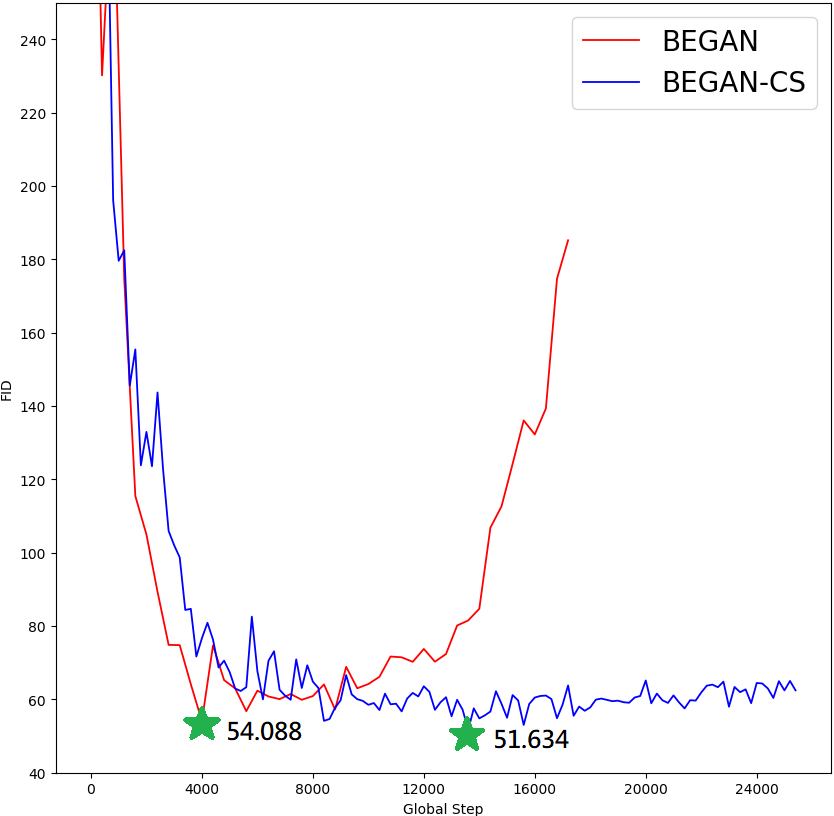}
        \end{tabular}
        \caption{FID through time. (Left) Full CelebA. (Right) 1/10 CelebA.}
        \label{Figure:FID_scores}
    \end{wrapfigure}

    For the quantitative comparison to demonstrate the effectiveness of the proposed constraint loss, we accordingly report \, ``Fréchet Inception Distance'' (FID)~\cite{HeuselRUNH17} score through time of BEGAN and BEGAN-CS in Fig.~\ref{Figure:FID_scores}. The experiments are conducted at $64\times 64$ resolution. It can be seen in Fig.~\ref{Figure:FID_scores} that, during training, the FID of BEGAN-CS does not increase drastically as BEGAN. 
    
    \subsection{Obtaining Optimal $z^*$ in One-Shot}
    
    In section \ref{subsection:obtain-z-in-one-shot}, we have shown that BEGAN-CS can approximate optimal $z^*$ with $Enc(x^*)$.  Appendix~A shows the experimental results of interpolation with obtained $z^*$ from $z^*$-search using different GAN architectures. The experiments may serve as proofs of concept for comparing the well-known GANs architectures, including FisherGAN~\cite{FisherGAN}, PGGAN~\cite{PGGAN}, and BEGAN. The experimental results show that the obtained $G(Enc(x^*))$ of BEGAN-CS is visually similar to $x^*$. In contrast, the original BEGAN and other state-of-the-art GANs require time-consuming $z^*$-search for 10{,}000 iterations to obtain competitive results. It would take 340 seconds to 3{,}970 seconds depending on the network architecture. However, the quality of the $z^*$-search result is still unstable and the searched image frequently looks quite different to the given real image, such as wrong gender or incorrect head pose. More examples on $z^*$-search with different GAN models and different numbers of optimization iterations are shown in Appendix~B.
    
    \subsection{Comparison with Bijective Models}
    \label{subsection:comparison-with-bijective}
    
\begin{figure}[!t]
        \centering
        \setlength{\tabcolsep}{4pt}
        \subfloat{\begin{tabular}[t]{ccccc}
            \includegraphics[width=0.15\linewidth]{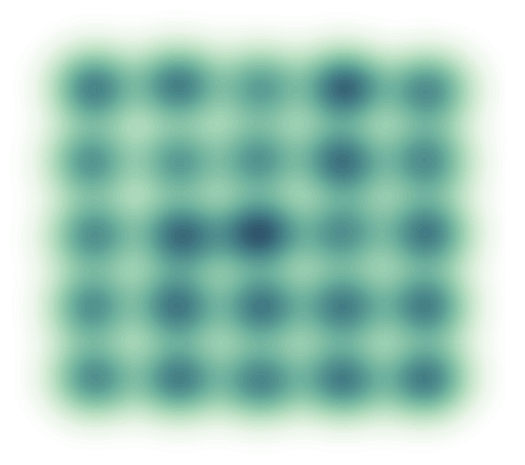} &
            \includegraphics[width=0.15\linewidth]{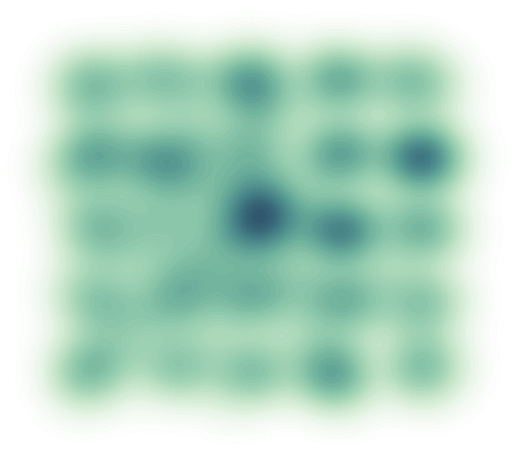} &
            \includegraphics[width=0.15\linewidth]{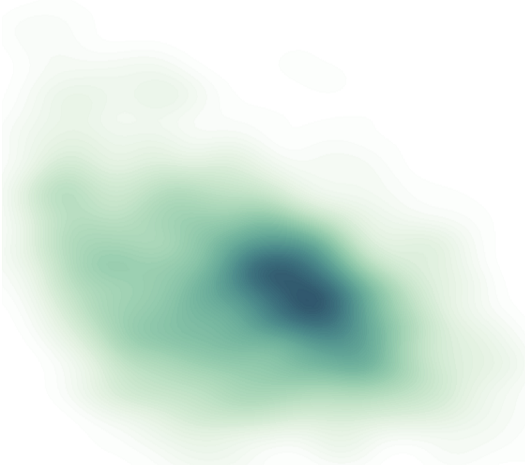}&
            \includegraphics[width=0.15\linewidth]{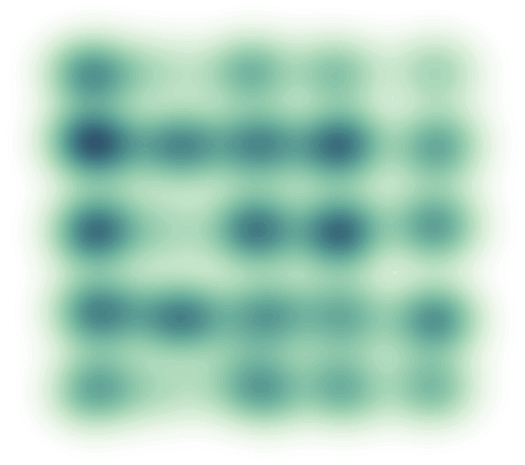} &
            \includegraphics[width=0.15\linewidth]{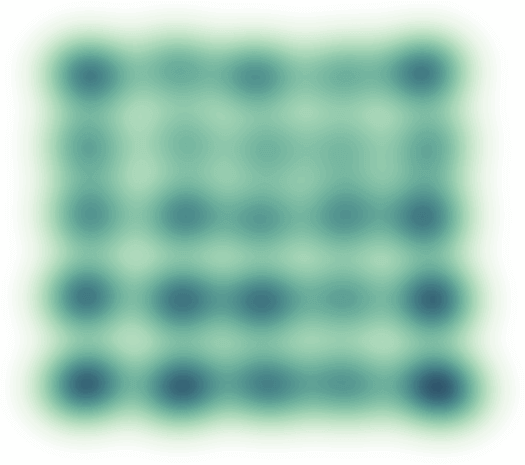} \\
            
            (a) Real &
            (b) VEEGAN &
            (c) ALI &
            (d) BEGAN &
            (e) BEGAN-CS
        \end{tabular}}
        \caption{Experimental results on the synthetic dataset introduced by VEEGAN.}
        \label{Figure:VEEGAN-Toy-Result}
        
    \end{figure}

    VEEGAN runs experiments on a synthetic toy dataset which consists of 25 independent Gaussian distributions, and observes better stable and higher diversity than other GANs. We accordingly run the similar experiment and provide comparisons in Fig.~\ref{Figure:VEEGAN-Toy-Result} for VEEGAN, ALI, BEGAN, and BEGAN-CS. We find that the vanilla BEGAN can already fit most of the modes of the real data distribution, though it requires extensive hyper-parameters tuning. Furthermore, BEGAN-CS can stabilize the training and converge to a final state of higher quality. Although VEEGAN can fit to all modes, the distribution is relatively blurry and less similar to the real data distribution. Lastly, ALI fails to fit to the real data distribution. 
    
    The hyper-parameters we used for BEGAN and BEGAN-CS on the toy dataset are $\alpha=0.1$, $\gamma$=25, $\lambda$=1e-4. We use Adam \cite{Adam} optimizer with ${lr}_d$=1e-4, ${lr}_g$=5e-4, $\beta_1$=0.5 and $\beta_2$=0.999. The latent dimension of $Z$ is set to 32. Both the generator and discriminator are consist of 2 layers of feed-forward network with 128 nodes and ReLU activation. We also set the weight initialization function to be a uniform-random sampler in range $[-\sqrt{9/n}, \sqrt{9/n}]$, which n is the number of layer input.
    
\begin{figure}[!t]
        \centering
        \setlength{\tabcolsep}{2pt}
        \subfloat{\begin{tabular}[t]{ccc}
            \includegraphics[width=0.22\linewidth]{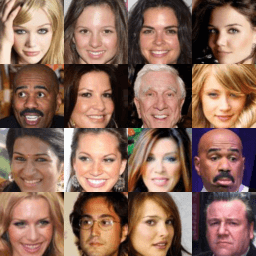} &
            \includegraphics[width=0.22\linewidth]{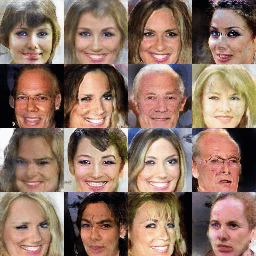} &
            \includegraphics[width=0.22\linewidth]{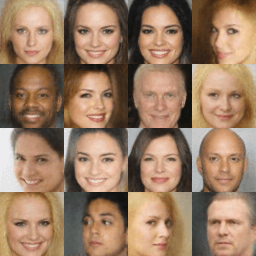} \\
            
            (a) Real images &
            (b) ALI &
            (c) BEGAN-CS \\
        \end{tabular}}
        \caption{Image reconstruction results.}
        \label{Figure:Compare_With_ALI}
    \end{figure}

    We also present qualitative comparisons on image reconstruction with BEGAN-CS and ALI in Fig.~\ref{Figure:Compare_With_ALI}. We find that the loss functions used by all three methods, ALI, BiGAN, and BEGAN-CS, do not guarantee that the reconstruction results are identical to the real images. BEGAN-CS is better at retaining some of the important features, such as hair color, skin color, gaze, and head pose.
    
    \subsection{On-the-Fly Representation Manipulation}

\begin{figure}[t]
        \centering
        \subfloat{\begin{tabular}[t]{c}
            \includegraphics[width=0.75\linewidth]{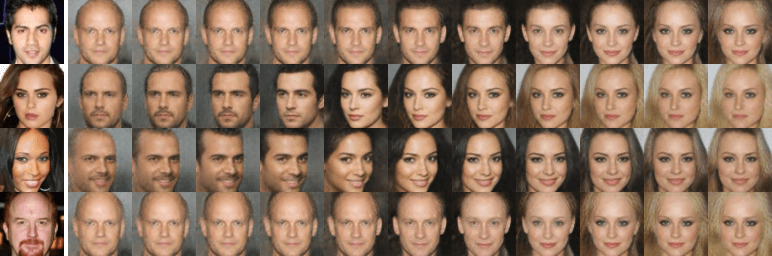} \\
                \small (a) Gender.
        \end{tabular}}
            
        \subfloat{\begin{tabular}[t]{c}
            \includegraphics[width=0.75\linewidth]{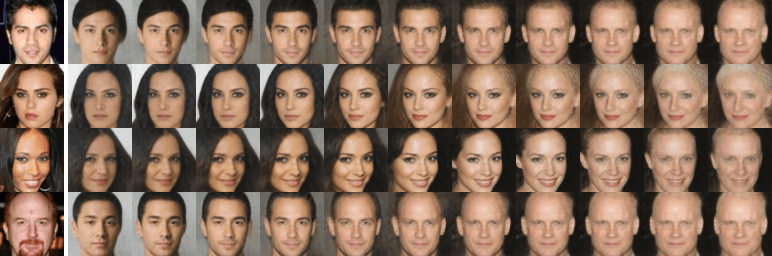} \\
                \small (b) Age.
        \end{tabular}}
    
        \subfloat{\begin{tabular}[t]{c}
            \includegraphics[width=0.75\linewidth]{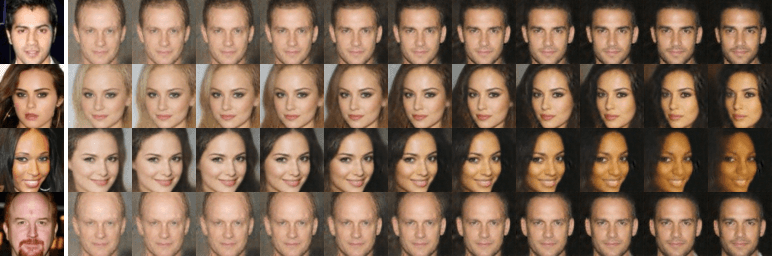} \\
                \small (c) Hair and skin color.
        \end{tabular}}
        
        \caption{Selected disentangled representations produced by BEGAN-CS at $64 \times 64$ resolution. For each series of images, the left-most image is the fixed real image $x^*$. In each sub-figure, we first obtain approximation of $z^*$ using $Enc(x^*)$. For each dimension $i$, we linearly interpolate and replace the $i$th dimension of $z^*$ by a value in $[-5, 5]$ with step size 1, and then generate the image set $\{G(z^*_i)\}$.}
        \label{figure:disentangled-representations}
    \end{figure}

\begin{figure}[b]
        \centering
        \null\hfill%
        \subfloat{\begin{tabular}[t]{c}
                \includegraphics[width=0.25\linewidth,height=0.25\linewidth]{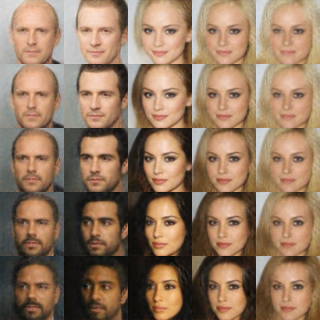} \\
                \begin{minipage}{0.25\linewidth}
                    \centering \small 
                    $\Leftrightarrow$: gender \\
                    $\Updownarrow$: hair and skin color
                \end{minipage}\\
            \end{tabular}}\hfill%
        \subfloat{\begin{tabular}[t]{c}
                \includegraphics[width=0.25\linewidth,height=0.25\linewidth]{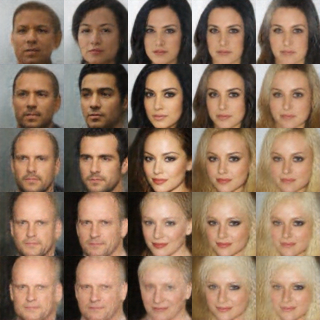} \\
                \begin{minipage}{0.25\linewidth}
                    \centering \small 
                    $\Leftrightarrow$: gender \\
                    $\Updownarrow$: age
                \end{minipage}\\
            \end{tabular}}\hfill%
        \subfloat{\begin{tabular}[t]{c}
                \includegraphics[width=0.25\linewidth,height=0.25\linewidth]{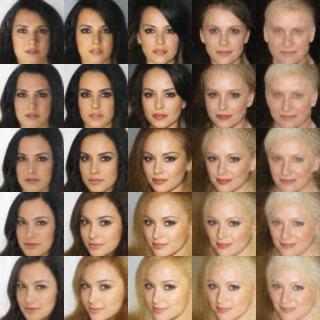} \\
                \begin{minipage}{0.25\linewidth}
                    \centering \small 
                    $\Leftrightarrow$: age \\
                    $\Updownarrow$: hairstyle
                \end{minipage}\\
            \end{tabular}}%
        \hfill\null%
        \caption{Two-dimensional combinations of disentangled representations.}
        \label{figure:adjust-multiple-disentangled-representations}
    \end{figure}

    In section \ref{subsection:method-disentangled-representation-and-application}, we demonstrate a new application of BEGAN-CS with the disentangled representations. By obtaining the approximation of $z^*$ with $Enc(x^*)$ and applying  the selected disentangled representations, BEGAN-CS can generate images that are visually similar to $x^*$ and exhibit the selected representations at the same time. As a proof of concept, we visualize the process of adding single representation in Fig.~\ref{figure:disentangled-representations} and multiple representations in Fig.~\ref{figure:adjust-multiple-disentangled-representations}.
    
    In Fig.~\ref{figure:disentangled-representations}, we first obtain the approximation of $z^*$ from $Enc(x^*)$. Then for each dimension $i$, we linearly interpolate and replace the value of latent vector $z^*$ at its $i$th dimension by a grid value in $[-5, 5]$ with step size 1, and thus can generate a series of images based on the modified latent vectors. The images show that each dimension of the latent space $Z$ represents a universal disentangled representation. We can perform similar visual transformations to any $z \in Z$. Fig.~\ref{figure:disentangled-representations} shows some of the interesting disentangled representations. The full visualization across the 64 dimensions is displayed in Appendix~C.
    
    The learned disentangled representations can also be used to perform multiple vector arithmetic operations on latent vectors. 
    This property enables us to control multiple attributes of a fixed image at the same time by adjusting multiple dimension values on the corresponding latent vector. We visualize the results of combining two different representations in Fig.~\ref{figure:adjust-multiple-disentangled-representations}.

\section{Conclusion}
    
    We identify that BEGAN suffers from the unpredictable mode-collapsing problem. The precise time when mode collapsing happens is non-deterministic, highly related to the resolution of generated images and the size of training dataset. We propose \emph{BEGAN with a Constrained Space} (BEGAN-CS) toward addressing the mode-collapsing problem and visualize the effect of constraint loss in the latent space. We experimentally show that the model-collapsing problem is suppressed after adding the constraint loss. BEGAN-CS performs particularly better than BEGAN when the size of training dataset is ten-times smaller than the normal setting. These advantages enable the class of energy-based GANs to move on to the next challenge of generating even higher resolution images.
    
    We also discover that BEGAN can learn salient and high-quality disentangled representations in an unsupervised setting. Combined with the particular property that BEGAN-CS is able to approximate $z^*$ on-the-fly, BEGAN-CS can generate images that are visually similar to the given real image and able to exhibit the adjustable disentangled properties. ``Obtaining $z^*$ in one-shot'' and ``adjustable image attributes'' are two interesting properties that have various potential applications, such as style manipulation and attribute-based editing.

    \clearpage
    

\begin{thebibliography}{10}

\bibitem{GAN}
Goodfellow, I.J., Pouget{-}Abadie, J., Mirza, M., Xu, B., Warde{-}Farley, D.,
  Ozair, S., Courville, A.C., Bengio, Y.:
\newblock Generative adversarial nets.
\newblock In: Advances in Neural Information Processing Systems 27: Annual
  Conference on Neural Information Processing Systems 2014, December 8-13 2014,
  Montreal, Quebec, Canada. (2014)  2672--2680

\bibitem{BousmalisSDEK17}
Bousmalis, K., Silberman, N., Dohan, D., Erhan, D., Krishnan, D.:
\newblock Unsupervised pixel-level domain adaptation with generative
  adversarial networks.
\newblock In: 2017 {IEEE} Conference on Computer Vision and Pattern
  Recognition, {CVPR} 2017, Honolulu, HI, USA, July 21-26, 2017. (2017)
  95--104

\bibitem{DaiFUL17}
Dai, B., Fidler, S., Urtasun, R., Lin, D.:
\newblock Towards diverse and natural image descriptions via a conditional
  {GAN}.
\newblock In: {IEEE} International Conference on Computer Vision, {ICCV} 2017,
  Venice, Italy, October 22-29, 2017. (2017)  2989--2998

\bibitem{GwakCGCS17}
Gwak, J., Choy, C.B., Garg, A., Chandraker, M., Savarese, S.:
\newblock Weakly supervised generative adversarial networks for 3d
  reconstruction.
\newblock CoRR \textbf{abs/1705.10904} (2017)

\bibitem{LedigTHCCAATTWS17}
Ledig, C., Theis, L., Huszar, F., Caballero, J., Cunningham, A., Acosta, A.,
  Aitken, A.P., Tejani, A., Totz, J., Wang, Z., Shi, W.:
\newblock Photo-realistic single image super-resolution using a generative
  adversarial network.
\newblock In: 2017 {IEEE} Conference on Computer Vision and Pattern
  Recognition, {CVPR} 2017, Honolulu, HI, USA, July 21-26, 2017. (2017)
  105--114

\bibitem{LiLY017}
Li, Y., Liu, S., Yang, J., Yang, M.:
\newblock Generative face completion.
\newblock In: 2017 {IEEE} Conference on Computer Vision and Pattern
  Recognition, {CVPR} 2017, Honolulu, HI, USA, July 21-26, 2017. (2017)
  5892--5900

\bibitem{ShrivastavaPTSW17}
Shrivastava, A., Pfister, T., Tuzel, O., Susskind, J., Wang, W., Webb, R.:
\newblock Learning from simulated and unsupervised images through adversarial
  training.
\newblock In: 2017 {IEEE} Conference on Computer Vision and Pattern
  Recognition, {CVPR} 2017, Honolulu, HI, USA, July 21-26, 2017. (2017)
  2242--2251

\bibitem{SoulySS17}
Souly, N., Spampinato, C., Shah, M.:
\newblock Semi supervised semantic segmentation using generative adversarial
  network.
\newblock In: {IEEE} International Conference on Computer Vision, {ICCV} 2017,
  Venice, Italy, October 22-29, 2017. (2017)  5689--5697

\bibitem{TzengHSD17}
Tzeng, E., Hoffman, J., Saenko, K., Darrell, T.:
\newblock Adversarial discriminative domain adaptation.
\newblock In: 2017 {IEEE} Conference on Computer Vision and Pattern
  Recognition, {CVPR} 2017, Honolulu, HI, USA, July 21-26, 2017. (2017)
  2962--2971

\bibitem{AutoEncoder}
Hinton, G.E., Salakhutdinov, R.R.:
\newblock Reducing the dimensionality of data with neural networks.
\newblock science \textbf{313}(5786) (2006)  504--507

\bibitem{BEGAN}
Berthelot, D., Schumm, T., Metz, L.:
\newblock {BEGAN:} boundary equilibrium generative adversarial networks.
\newblock CoRR \textbf{abs/1703.10717} (2017)

\bibitem{PCA}
:
\newblock Principal component analysis.
\newblock Chemometrics and Intelligent Laboratory Systems \textbf{2}(1)

\bibitem{DCGAN}
Radford, A., Metz, L., Chintala, S.:
\newblock Unsupervised representation learning with deep convolutional
  generative adversarial networks.
\newblock CoRR \textbf{abs/1511.06434} (2015)

\bibitem{SalimansGZCRCC16}
Salimans, T., Goodfellow, I.J., Zaremba, W., Cheung, V., Radford, A., Chen, X.:
\newblock Improved techniques for training gans.
\newblock In: Advances in Neural Information Processing Systems 29: Annual
  Conference on Neural Information Processing Systems 2016, December 5-10,
  2016, Barcelona, Spain. (2016)  2226--2234

\bibitem{EBGAN}
Zhao, J.J., Mathieu, M., LeCun, Y.:
\newblock Energy-based generative adversarial network.
\newblock CoRR \textbf{abs/1609.03126} (2016)

\bibitem{WGAN}
Arjovsky, M., Chintala, S., Bottou, L.:
\newblock Wasserstein {GAN}.
\newblock CoRR \textbf{abs/1701.07875} (2017)

\bibitem{PGGAN}
Karras, T., Aila, T., Laine, S., Lehtinen, J.:
\newblock Progressive growing of gans for improved quality, stability, and
  variation.
\newblock CoRR \textbf{abs/1710.10196} (2017)

\bibitem{DumoulinBPLAMC16}
Dumoulin, V., Belghazi, I., Poole, B., Lamb, A., Arjovsky, M., Mastropietro,
  O., Courville, A.C.:
\newblock Adversarially learned inference.
\newblock CoRR \textbf{abs/1606.00704} (2016)

\bibitem{DonahueKD16}
Donahue, J., Kr{\"{a}}henb{\"{u}}hl, P., Darrell, T.:
\newblock Adversarial feature learning.
\newblock CoRR \textbf{abs/1605.09782} (2016)

\bibitem{SrivastavaVRGS17}
Srivastava, A., Valkov, L., Russell, C., Gutmann, M.U., Sutton, C.A.:
\newblock {VEEGAN:} reducing mode collapse in gans using implicit variational
  learning.
\newblock In: Advances in Neural Information Processing Systems 30: Annual
  Conference on Neural Information Processing Systems 2017, 4-9 December 2017,
  Long Beach, CA, {USA}. (2017)  3310--3320

\bibitem{reviewer_paper_A}
Khan, S.H., Hayat, M., Barnes, N.:
\newblock Adversarial training of variational auto-encoders for high fidelity
  image generation.
\newblock In: 2018 {IEEE} Winter Conference on Applications of Computer Vision,
  {WACV} 2018, Lake Tahoe, NV, USA, March 12-15, 2018. (2018)  1312--1320

\bibitem{t-SNE}
Maaten, L.v.d., Hinton, G.:
\newblock Visualizing data using t-sne.
\newblock Journal of machine learning research \textbf{9}(Nov) (2008)
  2579--2605

\bibitem{CelebA}
Liu, Z., Luo, P., Wang, X., Tang, X.:
\newblock Deep learning face attributes in the wild.
\newblock In: Proceedings of International Conference on Computer Vision
  (ICCV). (2015)

\bibitem{VAE}
Kingma, D.P., Welling, M.:
\newblock Auto-encoding variational bayes.
\newblock CoRR \textbf{abs/1312.6114} (2013)

\bibitem{beta-VAE}
Higgins, I., Matthey, L., Pal, A., Burgess, C., Glorot, X., Botvinick, M.,
  Mohamed, S., Lerchner, A.:
\newblock beta-vae: Learning basic visual concepts with a constrained
  variational framework.
\newblock (2016)

\bibitem{VAE-GAN}
Larsen, A.B.L., S{\o}nderby, S.K., Larochelle, H., Winther, O.:
\newblock Autoencoding beyond pixels using a learned similarity metric.
\newblock In: Proceedings of the 33nd International Conference on Machine
  Learning, {ICML} 2016, New York City, NY, USA, June 19-24, 2016. (2016)
  1558--1566

\bibitem{info-GANs}
Chen, X., Chen, X., Duan, Y., Houthooft, R., Schulman, J., Sutskever, I.,
  Abbeel, P.:
\newblock Infogan: Interpretable representation learning by information
  maximizing generative adversarial nets.
\newblock In: Advances in Neural Information Processing Systems 29: Annual
  Conference on Neural Information Processing Systems 2016, December 5-10,
  2016, Barcelona, Spain. (2016)  2172--2180

\bibitem{HeuselRUNH17}
Heusel, M., Ramsauer, H., Unterthiner, T., Nessler, B., Hochreiter, S.:
\newblock Gans trained by a two time-scale update rule converge to a local nash
  equilibrium.
\newblock In: Advances in Neural Information Processing Systems 30: Annual
  Conference on Neural Information Processing Systems 2017, 4-9 December 2017,
  Long Beach, CA, {USA}. (2017)  6629--6640

\bibitem{FisherGAN}
Mroueh, Y., Sercu, T.:
\newblock Fisher {GAN}.
\newblock In: Advances in Neural Information Processing Systems 30: Annual
  Conference on Neural Information Processing Systems 2017, 4-9 December 2017,
  Long Beach, CA, {USA}. (2017)  2510--2520

\bibitem{Adam}
Kingma, D.P., Ba, J.:
\newblock Adam: {A} method for stochastic optimization.
\newblock CoRR \textbf{abs/1412.6980} (2014)

\bibitem{SN}
Miyato, T., Kataoka, T., Koyama, M., Yoshida, Y.:
\newblock Spectral normalization for generative adversarial networks.
\newblock CoRR \textbf{abs/1802.05957} (2018)

\end{thebibliography}

    \clearpage
    
    \setcounter{secnumdepth}{0}
    
    \section{Appendix A: Interpolation and $z^*$-Search}
    
    \vfill
    
    \begin{figure}[!h]
        \captionsetup[subfigure]{justification=centering}
        \centering
        \subfloat{\begin{tabular}[t]{c}
            \includegraphics[width=0.92\linewidth]{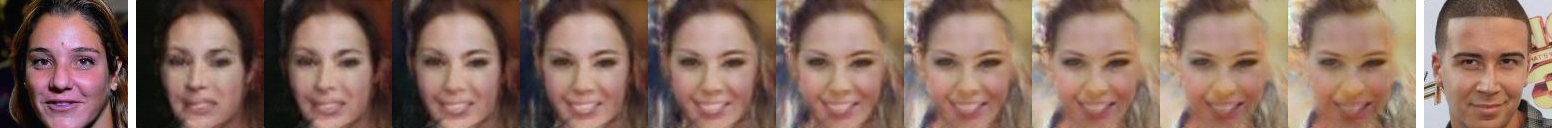}\\
            \small (a) FisherGAN \cite{FisherGAN} + Spectral Normalization \cite{SN} ($128 \times 128$) \\
            \small ($z^*$-search 10,000 iterations, taking 340 seconds)
        \end{tabular}}
        
        \centering
        \subfloat{\begin{tabular}[t]{c}
            \includegraphics[width=0.92\linewidth]{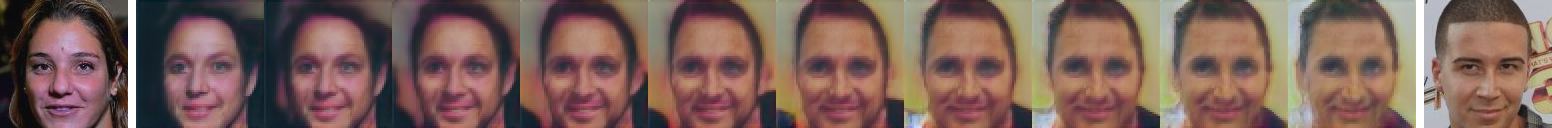} \\
                \small (b) PGGANs \cite{PGGAN} ($128 \times 128$) \\
                \small ($z^*$-search 10,000 iterations, taking 710 seconds)
        \end{tabular}}
        
        \centering
        \subfloat{\begin{tabular}[t]{c}
            \includegraphics[width=0.92\linewidth]{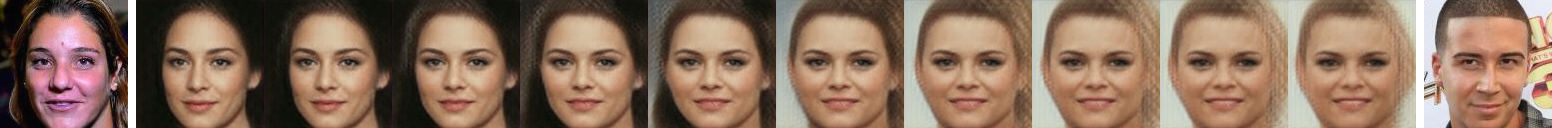} \\
                \small (c) BEGAN ($128 \times 128$) \\
                \small ($z^*$-search 10,000 iterations, taking 4{,}300 seconds)
        \end{tabular}}
        
        \centering
        \subfloat{\begin{tabular}[t]{c}
            \includegraphics[width=0.92\linewidth]{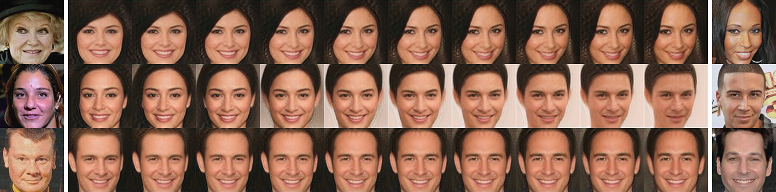} \\
                \small (d) BEGAN-CS ($128 \times 128$) \\
                \small (without $z^*$-search, on-the-fly)
        \end{tabular}}
        
        \caption{Interpolation between two real images in latent space using different GAN models. Other state-of-the-art GANs require time-consuming $z^*$-search for 10{,}000 iterations to obtain competitive results, taking several minutes. Nevertheless, the quality of the $z^*$-search results is still not as good as the quality of the images generated  on-the-fly by BEGAN-CS. }
        \label{figure:z-search-in-one-shot}
    \end{figure}
    
    \vfill
    
    \clearpage
    
    \section{Appendix B: More $z^*$-Search}
    
    \vfill
    
    \begin{figure}[!h]
        \begin{tabular}{L{.16\linewidth}L{0.42\linewidth}L{0.42\linewidth}}
            \begin{tabular}{c}
                \scriptsize Real images $x^*$ \\ [0.9em]
                \scriptsize $G(Enc(x^*))$ \\ [0.9em]
                \scriptsize 100 iters \\ [0.9em]
                \scriptsize 500 iters \\ [0.9em]
                \scriptsize 1{,}000 iters \\ [0.9em]
                \scriptsize 3{,}000 iters \\ [0.9em]
                \scriptsize 5{,}000 iters \\ [0.9em]
                \scriptsize 10{,}000 iters
                \end{tabular} &
            \subfloat{\begin{tabular}[t]{c}
                \includegraphics[width=\linewidth]{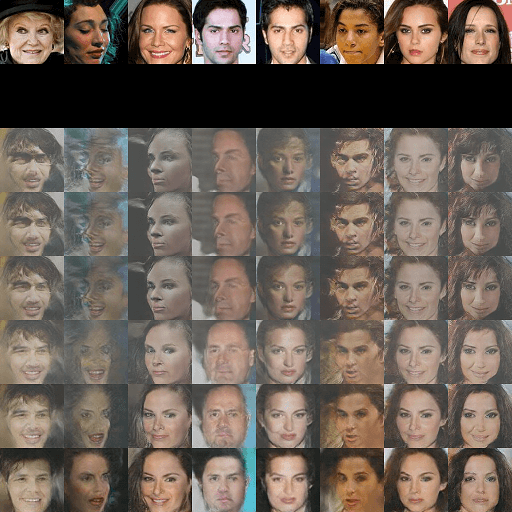} \\
                \small (a) FisherGAN+Spectral Norm
            \end{tabular}} &
            \subfloat{\begin{tabular}[t]{c}
                \includegraphics[width=\linewidth]{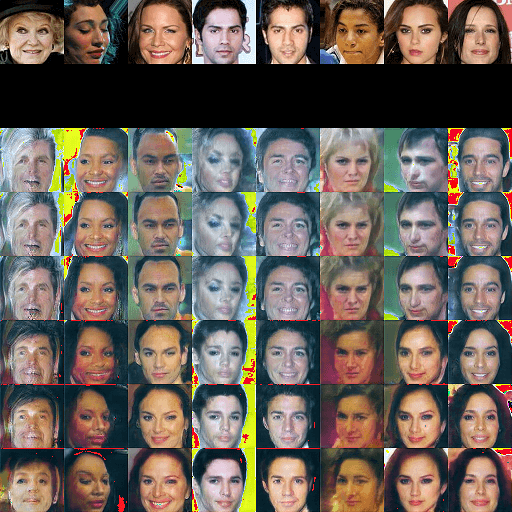} \\
                \small (b) PGGAN
            \end{tabular}}
        \end{tabular}
        
        \begin{tabular}{L{.16\linewidth}L{0.42\linewidth}L{0.42\linewidth}}
            \begin{tabular}{c}
                \scriptsize Real images $x^*$ \\ [0.9em]
                \scriptsize $G(Enc(x^*))$ \\ [0.9em]
                \scriptsize 100 iters \\ [0.9em]
                \scriptsize 500 iters \\ [0.9em]
                \scriptsize 1{,}000 iters \\ [0.9em]
                \scriptsize 3{,}000 iters \\ [0.9em]
                \scriptsize 5{,}000 iters \\ [0.9em]
                \scriptsize 10{,}000 iters
                \end{tabular} &
            \subfloat{\begin{tabular}[t]{c}
                \includegraphics[width=\linewidth]{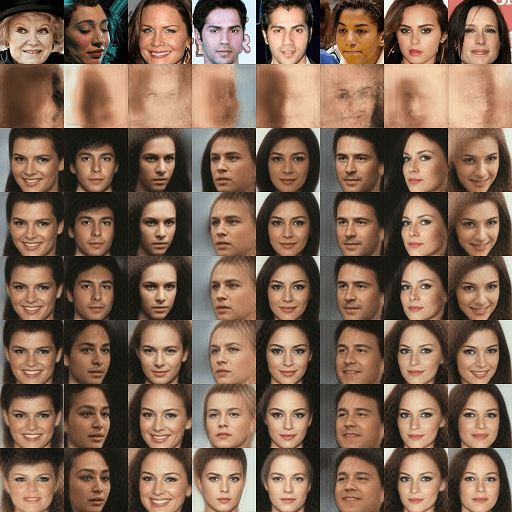} \\
                \small (c) BEGAN
            \end{tabular}} &
            \subfloat{\begin{tabular}[t]{c}
                \includegraphics[width=\linewidth]{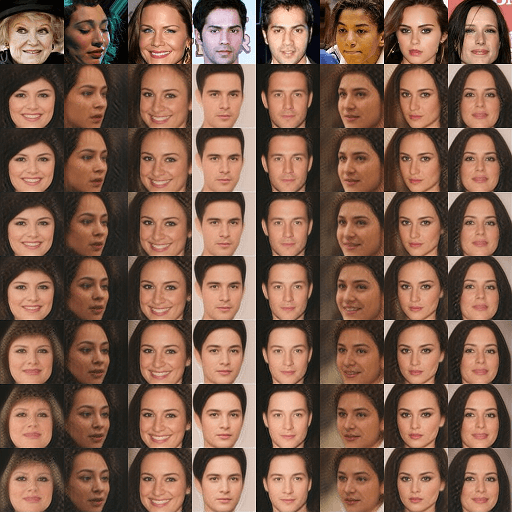} \\
            \small (d) BEGAN-CS
            \end{tabular}}
        \end{tabular}
        
        \caption{(a-c) We perform $z^*$-search from a random starting point $z \in Z$ for FisherGAN, PGGAN, and BEGAN. (d) BEGAN-CS starts from $Enc(x)$. We also show the result of $G(Enc(x^*))$ for BEGAN and BEGAN-CS. It can be seen that only for BEGAN-CS the result of $G(Enc(x^*))$ can be considered as a good approximation of $z^*$.}
        \label{fig:more-z-search}
    \end{figure}
    
    \vfill
    
    \clearpage
    
    \section{Appendix C: All Disentangled Representations}
    
    \begin{figure}[H]
        \centering
        \begin{tabular}{C{0.5\linewidth}C{0.5\linewidth}}
             \subfloat{\includegraphics[width=0.95\linewidth]{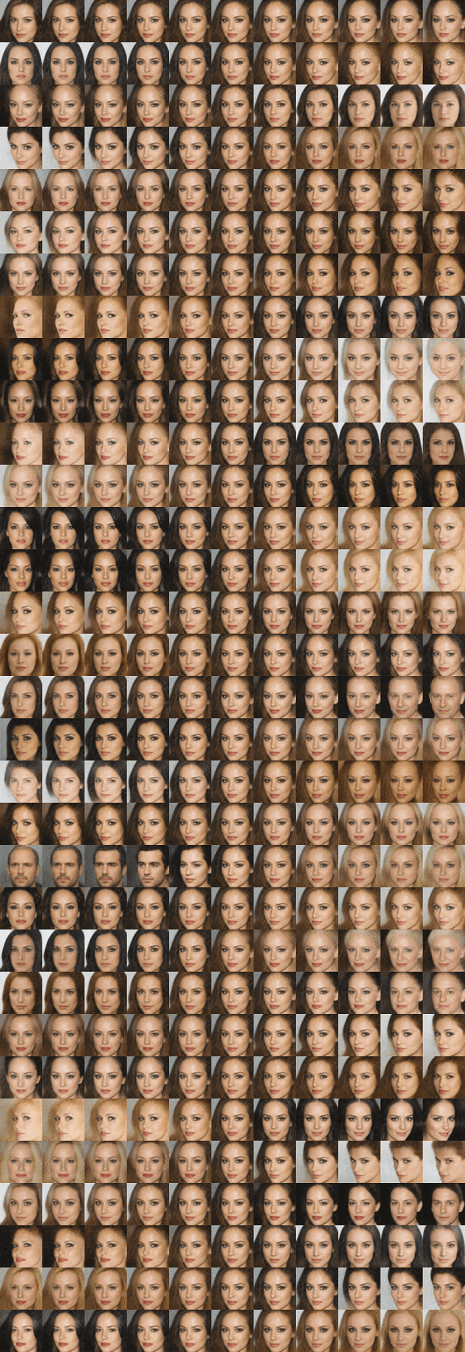}} &
             \subfloat{\includegraphics[width=0.95\linewidth]{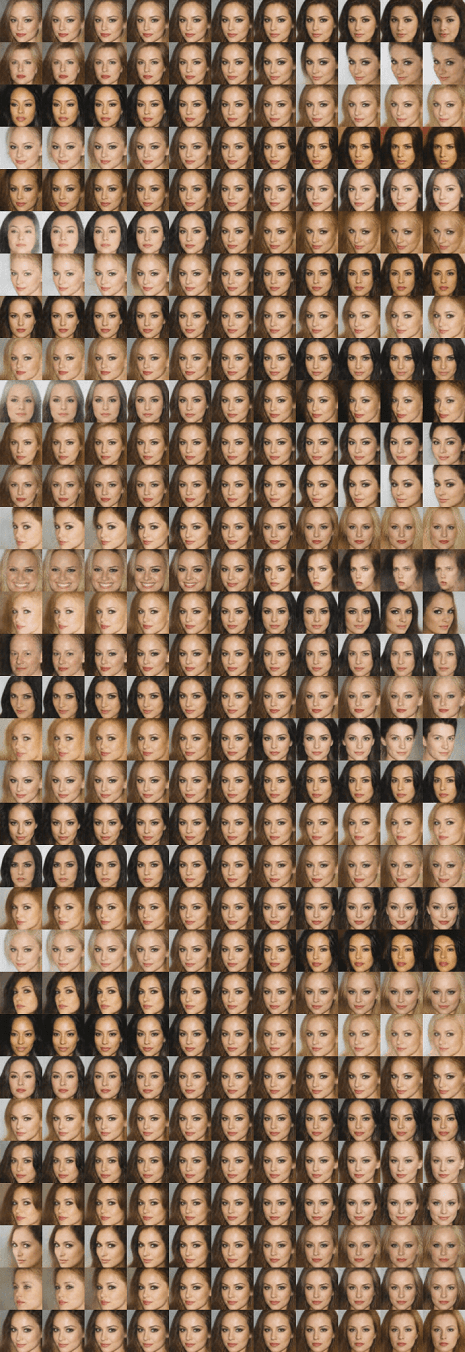}}
        \end{tabular}
        \caption{Disentangled representations of BEGAN-CS across 64 dimensions along each axis in latent space $Z$.
        \label{figure:all-disentangled-representations}}
    \end{figure}

\end{document}